\documentclass[10pt,twocolumn,letterpaper]{article}

\usepackage{cvpr}
\usepackage{placeins}
\usepackage{times}
\usepackage{epsfig}
\usepackage{graphicx}
\usepackage{amsmath}
\usepackage{amssymb}
\usepackage{textcomp}

\usepackage{eqnarray}
\usepackage{amsmath}
\usepackage{units}
\usepackage{graphicx}
\usepackage{caption}
\usepackage{subcaption}
\usepackage{url}
\usepackage{algorithmicx}
\usepackage{algpseudocode}
\usepackage{mathtools}
\usepackage{bbm}
\usepackage{stackengine}
\usepackage{xfrac}

\usepackage{multirow}
\usepackage{colortbl}
\usepackage{appendix}

\allowdisplaybreaks

\DeclarePairedDelimiter{\abs}{\lvert}{\rvert}
\DeclarePairedDelimiter{\norm}{\lVert}{\rVert}
\DeclareMathOperator{\atantwo}{atan2}
\DeclareMathOperator{\diag}{diag}

\definecolor{Yellow}{rgb}{1,1, 0.6}
\definecolor{Red}{rgb}{1, 0.6, 0.6}

\newcommand{\floor}[1]{\left \lfloor #1 \right \rfloor}
\newcommand{\binwidth}{h}
\newcommand{\E}[1]{\operatorname{E}\left\lbrack#1\right\rbrack}
\newcommand{\fftv}[1]{\mathcal{F}_\mathrm{v} \left( #1 \right)}
\newcommand{\ifftv}[1]{\mathcal{F}^{-1}_\mathrm{v} \left( #1 \right)}
\newcommand{\fft}[1]{\mathcal{F}\left( #1 \right )}

\newcommand{\lossdata}[1]{f\left( #1 \right )}
\DeclareMathOperator*{\argmax}{arg\,max}
\DeclareMathOperator*{\argmin}{arg\,min}

\newcommand{\sixwidth}{1.103in}
\newcommand{\threewidth}{2.22in}
\newcommand{\ishift}{\bar{i}}
\newcommand{\jshift}{\bar{j}}

\usepackage{expl3}
\ExplSyntaxOn
\newcommand\latinabbrev[1]{
  \peek_meaning:NTF . {
    #1\@}%
  { \peek_catcode:NTF a {
      #1.\@ }%
    {#1.\@}}}
\ExplSyntaxOff
\def\etal{\latinabbrev{et al}}

\usepackage[breaklinks=true,letterpaper=true,bookmarks=false]{hyperref}

\cvprfinalcopy 

\def\cvprPaperID{287} 

\ifcvprfinal\fi
\begin{document}

\title{Fast Fourier Color Constancy}

\author{
Jonathan T. Barron\\
{\tt\small barron@google.com}
\and
Yun-Ta Tsai\\
{\tt\small yuntatsai@google.com}
}
\maketitle

\begin{abstract}
We present Fast Fourier Color Constancy (FFCC), a color constancy algorithm
which solves illuminant estimation by reducing it to a spatial localization
task on a torus.
By operating in the frequency domain,
FFCC produces lower error rates than the previous state-of-the-art by
$13-20\%$ while being $250-3000\times$ faster.
This unconventional approach introduces challenges regarding aliasing,
directional statistics, and preconditioning, which we address.
By producing a complete posterior distribution over illuminants instead of a
single illuminant estimate, FFCC enables better training techniques,
an effective temporal smoothing technique,
and richer methods for error analysis.
Our implementation of FFCC runs at $\sim 700$ frames per second on a mobile
device, allowing it to be used as an accurate,
real-time, temporally-coherent automatic white balance algorithm.
\end{abstract}

\section{Intro}

A fundamental problem in computer vision is that of estimating the underlying
world that resulted in some observed image \cite{Adelson1996, BarrowTenenbaum78}.
One subset of this problem is color constancy:
estimating the color of the illuminant of the scene and the colors of the objects in the
scene viewed under a white light.
Despite its apparent simplicity, this problem has yielded a great deal
of depth and challenge for both the human vision and computer vision communities
\cite{foster2011color, GijsenijTIP2011}.
Color constancy is also a practical concern in the camera industry:
producing a natural looking photograph without user intervention requires that
the illuminant be automatically estimated and discounted, a process referred to as
``auto white balance'' among practitioners.
Though there is a profound historical connection between color constancy and
consumer photography
(exemplified by Edwin Land, the inventor of both Retinex theory
\cite{land1971lightness} and the Polaroid instant camera),
``color constancy'' and ``white balance'' have come to mean different things
--- color constancy aims to recover the veridical world behind an image,
while white balance aims to give an image a pleasant appearance
consistent with some aesthetic or cultural norm.
But with the current ubiquity of learning-based techniques in computer vision,
both problems reduce to just estimating the
``best'' illuminant from an image, and the question of whether that illuminant
is objectively true or subjectively attractive is just a matter of the
data used during training.

Despite their accuracy, modern learning-based color constancy algorithms are
not immediately suitable as practical white balance algorithms, as practical
white balance has several requirements besides accuracy:
\\ {\bf Speed} -
An algorithm running in a camera's viewfinder must run at $30$ FPS
on mobile hardware.
But a camera's compute budget is precious: demosaicing, face detection,
auto exposure, etc, must also run simultaneously and in real time.
Spending more than a small fraction (say, $5-10\%$) of a camera's
compute budget on white balance is impractical, suggesting that our speed
requirement is closer to $1-5$ milliseconds per frame.
\\ {\bf Impoverished Input} -
Most color constancy algorithms are designed for full resolution, high
bit-depth input images, but operating on such large images is challenging and
costly in practice.
To be fast, the algorithm must work well on the small, low bit-depth
``preview'' images ($32 \times 24$ or $64 \times 48$ pixels, $8$-bit)
which are usually computed by specialized camera hardware for this task.
\\ {\bf Uncertainty} -
In addition to the illuminant, the algorithm should produce
some confidence measure or a complete posterior distribution over
illuminants, thereby enabling convenient downstream integration with
hand-engineered heuristics or external sources of information.
\\ {\bf Temporal Coherence} -
The algorithm should allow the estimated illuminant to be smoothed over time,
to prevent color composition in videos from varying erratically.

In this paper we present a novel color constancy algorithm, which we call
``Fast Fourier Color Constancy'' (FFCC).
Viewed as a color constancy algorithm, FFCC is $13-20\%$ more accurate than the
state of the art on standard benchmarks.
Viewed as a prospective white balance algorithm, FFCC addresses our
previously described requirements: Our technique is
$250-3000\times$ faster than the state of the art, and is capable of running at
$1.44$ milliseconds per frame on a standard consumer mobile platform using the
thumbnail images already produced by that camera's hardware.
FFCC produces a complete posterior distribution over illuminants which allows
us to reason about uncertainty and enables simple and effective temporal smoothing.

We build on the ``Convolutional Color Constancy'' (CCC) approach
of \cite{BarronICCV2015}, which is
currently one of the top-performing techniques on standard color constancy
benchmarks \cite{Cheng14,Gehler08,shifunt}.
CCC works by observing that applying a per-channel gain to a linear RGB
image is equivalent to inducing a 2D translation of the log-chroma histogram
of that image, which allows color constancy to be reduced to the
task of localizing a signature in log-chroma histogram space.
This reduction is at the core of the success of CCC and, by extension,
our FFCC technique;
see \cite{BarronICCV2015} for a thorough explanation.
The primary difference between FFCC is that instead of performing an expensive
localization on a large log-chroma plane, we perform a cheap
localization on a small log-chroma \emph{torus}.

\begin{figure}[!]
\centering
  \stackunder[5pt]
    {\includegraphics[width=1.55in]{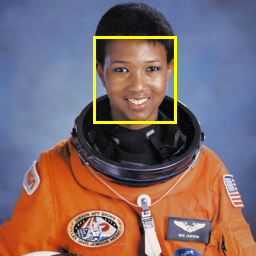}}
    {(a) Image $A$}
  \stackunder[5pt]
    {\includegraphics[width=1.55in]{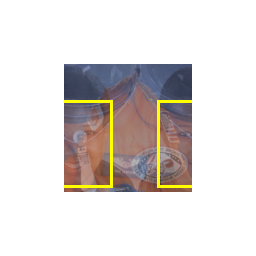}}
    {(b) Aliased Image $B$}
  \caption{
  CCC \cite{BarronICCV2015} reduces color constancy
  to a 2D localization problem similar to object detection (1a).
  FFCC repeatedly wraps this 2D localization problem around a small torus (1b),
  which creates challenges but allows for faster illuminant estimation.
  See the text for details.
  \label{fig:face}
  }
\end{figure}

\newcommand{\mypm}{\mathbin{\smash{%
\raisebox{0.35ex}{%
            $\underset{\raisebox{0.5ex}{$\smash -$}}{\smash+}$%
            }%
        }%
    }%
}

At a high level, CCC reduces color constancy to object detection ---
in the computability theory sense of ``reduce''.
FFCC reduces color constancy to localization on a torus instead of a plane,
and because this task has no intuitive analogue in computer vision we will
attempt to provide one\footnote{
We cannot speak to the merit of this idea in the context of object detection,
and we present it here solely to provide an intuition of our work on color
constancy}.
Given a large image $A$ on which we would like to perform object detection,
imagine constructing a smaller $n \times n$ image $B$ in which each pixel in $B$
is the sum of all values in $A$ separated by a multiple of $n$ pixels in either
dimension:
\begin{equation}
B(i,j) = \sum_{k, l} A(i + nk, j + nl)
\end{equation}
This amounts to taking $A$ and repeatedly wrapping it around a small torus
(see Figure~\ref{fig:face}).
Detecting objects this way may yield a speedup as the image being
processed is smaller, but it also raises new problems:
1) pixel values are corrupted with superimposed shapes that make detection difficult,
2) detections must ``wrap'' around the edges of this toroidal image,
and 3) instead of an absolute, global location we can only recover an aliased,
incomplete location.
FFCC works by taking the large convolutional problem of CCC
(ie, face detection on $A$)
and aliasing that problem down to a smaller size where it can be solved efficiently
(ie, face detection on $B$).
We will show that we can learn an effective color constancy model in the face of
the difficulty and ambiguity introduced by aliasing.
This convolutional classifier will be implemented and learned using FFTs,
because the naturally periodic nature of FFT convolutions resolves
the problem of detections ``wrapping'' around the edge of toroidal images,
and produces a significant speedup.

Our approach to color constancy introduces a number of issues.
The aforementioned periodic ambiguity resulting from operating
on a torus (which we dub ``illuminant aliasing'') requires new techniques
for recovering a global illuminant estimate from an aliased estimate
(Section~\ref{sec:aliasing}).
Localizing the centroid of the illuminant on a torus is difficult, requiring
that we adopt and extend techniques from the directional statistics literature
(Section~\ref{sec:bvm}).
But our approach presents a number of benefits.
FFCC improves accuracy relative to CCC by $17-24\%$ while retaining its flexibility,
and allows us to construct priors over illuminants (Section~\ref{sec:extensions}).
By learning in the frequency-domain we can
construct a novel method for fast frequency-domain regularization and
preconditioning, making FFCC training $20\times$ faster than CCC
(Section~\ref{sec:fourier}).
Our model produces a complete unimodal posterior over illuminants as output,
allowing us to construct a Kalman filter-like approach for processing videos
instead of independent images
(Section~\ref{sec:temporal}).

\begin{figure*}[!]
\centering
  \stackunder[5pt]
    {\includegraphics[width=\sixwidth]{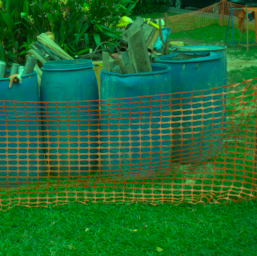}}
    {\footnotesize (a) Input Image}
  \stackunder[5pt]
    {\includegraphics[width=\sixwidth]{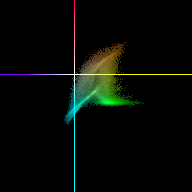}}
    {\footnotesize (b) Histogram}
  \stackunder[5pt]
    {\includegraphics[width=\sixwidth]{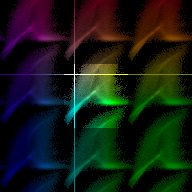}}
    {\footnotesize (c) Aliased Histogram}
  \stackunder[5pt]
    {\includegraphics[width=\sixwidth]{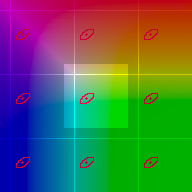}}
    {\footnotesize (d) Aliased Prediction}
  \stackunder[5pt]
    {\includegraphics[width=\sixwidth]{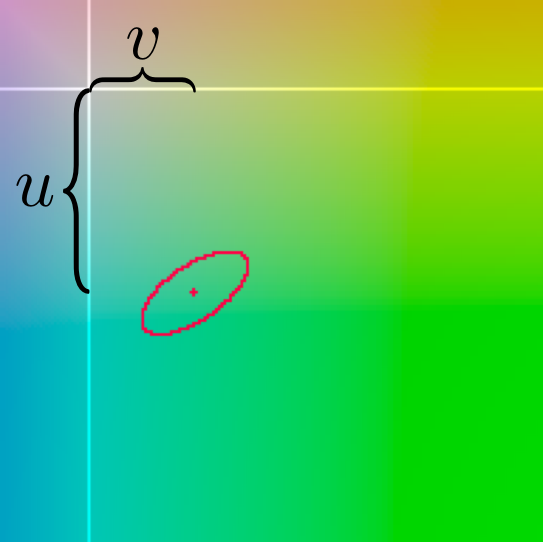}}
    {\footnotesize (e) De-aliased Prediction}
  \stackunder[5pt]
    {\includegraphics[width=\sixwidth]{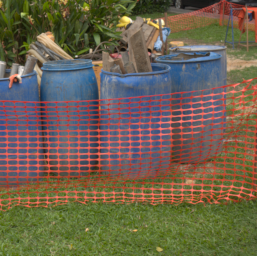}}
    {\footnotesize (f) Output Image}
    \caption{
    An overview of our pipeline demonstrating the problem of illuminant
    aliasing.
    Similarly to CCC, we take an input image (2a) and
    transform it into a log-chroma histogram (2b, presented
    in the same format as in \cite{BarronICCV2015}).
    But unlike CCC, our histograms are small and toroidal, meaning that pixels
    can ``wrap around'' the edges (2c, with the torus
    ``unwrapped'' once in every direction). This means that the
    centroid of a filtered histogram, which would simply be \emph{the} illuminant
    estimate in CCC, is instead an infinite family of possible illuminants (2d).
    This requires \emph{de-aliasing}, some technique for disambiguating between
    illuminants to select the single most likely estimate
    (2e, shown as a point surrounded by an ellipse visualizing
    the output covariance of our model).
    Our model's output $(u, v)$ coordinates in this de-aliased log-chroma space
    corresponds to the color of the illuminant, which can then be divided
    into the input image to produce a white balanced image (2f).
    \label{fig:overview}
    }
\end{figure*}

\section{Convolutional Color Constancy}
\label{sec:ccc}

Let us review the assumptions made in CCC and inherited by our model.
Assume that we have a photometrically linear input image $I$ from a camera,
with a black level of zero and with no saturated
pixels\footnote{in practice, saturated pixels are identified and removed from all
downstream computation, similarly to how color checker pixels are ignored.}.
Each pixel $k$'s RGB value in image $I$ is assumed to be the product of that
pixel's ``true'' white-balanced RGB value $W^{(k)}$
and some global RGB illumination $L$ shared by all pixels:
\begin{align}
\forall_k \,\,
\begin{bmatrix} I_r^{(k)} \\ I_g^{(k)} \\ I_b^{(k)} \end{bmatrix} = \begin{bmatrix} W_r^{(k)} \\ W_g^{(k)} \\ W_b^{(k)} \end{bmatrix} \circ \begin{bmatrix} L_r \\ L_g \\ L_b \end{bmatrix}
\label{eq:image_formation_rgb}
\end{align}
The task of color constancy is to use the input image $I$ to estimate $L$, and
with that produce $W^{(k)} = I^{(k)} / L$.

Given a pixel from our input RGB image $I^{(k)}$, CCC defines two log-chroma
measures:
\begin{align}
u^{(k)} &= \log \left( I^{(k)}_g / I^{(k)}_r \right) & v^{(k)} &= \log \left( I^{(k)}_g / I^{(k)}_b \right)
\end{align}
The absolute scale of $L$ is assumed to be unrecoverable,
so estimating $L$ simply requires estimating its log-chroma:
\begin{align}
L_u &= \log \left( L_g / L_r \right) & L_v &= \log \left( L_g / L_b \right)
\end{align}
After recovering $(L_u, L_v)$, assuming that $L$ has a magnitude of $1$ lets us
recover the RGB values of the illuminant:
\begin{align}
L_r &= {\exp(-L_u) \over z} \qquad L_g = {1 \over z} \qquad L_b = {\exp(-L_v) \over z} \nonumber \\
z &= \sqrt{\exp(-L_u)^2 + \exp(-L_v)^2 + 1}  \label{eq:light_rgb}
\end{align}
Framing color constancy in terms of predicting log-chroma has several small
advantages over the standard RGB approach ($2$ unknowns instead of $3$, better
numerical stability, etc) but the primary advantage of this approach is that using
log-chroma turns the multiplicative constraint relating $W$ and $I$ into
an additive constraint \cite{finlayson2001color}, and this in turn enables a
convolutional approach to color constancy.
As shown in \cite{BarronICCV2015}, color constancy can be framed as a 2D
spatial localization task on a log-chroma histogram $N$, where some
sliding-window classifier is used to filter that histogram and the centroid of
that filtered histogram is used as the log-chroma of the illuminant.

\section{Illuminant Aliasing}
\label{sec:aliasing}

We assume the same convolutional premise of CCC, but with one primary
difference to improve quality and speed: we use FFTs to perform the convolution
that filters the log-chroma histogram, and we use a small histogram
to make that convolution as fast as possible. This change may seem trivial,
but the periodic nature of FFT convolution combined with the properties of
natural images has a significant effect, as we will demonstrate.

Similarly to CCC, given an input image $I$ we construct a histogram $N$ from $I$,
where $N(i,j)$ is the number of pixels in $I$ whose log-chroma is near the
$(u, v)$ coordinates corresponding to histogram position $(i, j)$:
\begin{align}
N(i, j) = \displaystyle \sum_k \Bigg( & \operatorname{mod} \left( { u^{(k)} - u_\mathit{lo} \over \binwidth} - i, n \right) < 1 \nonumber \\
\wedge \,\, & \operatorname{mod} \left( { v^{(k)} - v_\mathit{lo} \over \binwidth} - j, n \right) < 1 \Bigg) \label{eq:Nhist}
\end{align}
Where $i$, $j$ are $0$-indexed,
$n = 64$ is the number of bins,
$\binwidth = 1/32$ is the bin size, and
$(u_{\mathit{lo}}, v_{\mathit{lo}})$ is the starting point of the histogram.
Because our histogram is too small to contain the wide spread of colors present
in most natural images,
we use modular arithmetic to cause pixels to ``wrap around'' with respect to log-chroma
(any other standard boundary condition would violate our convolutional
assumption and would cause many image pixels to be ignored).
This means that, unlike standard CCC,
a single $(i,j)$ coordinate in the histogram no longer corresponds to
an absolute $(u,v)$ color, but instead corresponds to an infinite family of
$(u,v)$ colors.
Accordingly, the
centroid of a filtered histogram no longer corresponds to the color of the
illuminant, but instead is an infinite set of illuminants.
We will refer to this phenomenon as \emph{illuminant aliasing}.
Solving this problem requires that we use some technique to de-alias an aliased
illuminant estimate\footnote{It is tempting to refer to resolving the illuminant
aliasing problem as ``anti-aliasing'', but anti-aliasing usually refers to
preprocessing a signal to prevent aliasing during some resampling operation,
which does not appear possible in our framework.
``De-aliasing'' suggests that we allow aliasing to happen to the input, but then
remove the aliasing from the output.}.
A high-level outline of our FFCC pipeline that illustrates illuminant (de-)aliasing
can be seen in Fig.~\ref{fig:overview}.

De-aliasing requires that we use some external information (or some external
color constancy algorithm) to disambiguate between illuminants.
An intuitive approach is to select the illuminant
that causes the average image color to be as neutral as possible,
which we call ``gray world de-aliasing''.
We compute average log-chroma values $(\bar{u}, \bar{v})$ for the entire image
and use this to turn an aliased illuminant estimate $(\hat{L}_u, \hat{L}_v)$
into a de-aliased illuminant $(\hat{L}'_u, \hat{L}'_v)$:
\begin{align}
\bar{u} = \operatorname{mean}_k \left( u^{(k)} \right) \quad\quad
\bar{v} = \operatorname{mean}_k \left( v^{(k)} \right) \\
\begin{bmatrix} \hat{L}'_u \\ \hat{L}'_v \end{bmatrix} = \begin{bmatrix} \hat{L}_u \\ \hat{L}_v \end{bmatrix} - (n \binwidth) \floor{{1 \over n \binwidth} \begin{bmatrix} \hat{L}_u - \bar{u} \\ \hat{L}_v - \bar{v} \end{bmatrix} + {1 \over 2} } 
\label{eq:grayworld}
\end{align}
Another approach, which we call ``gray light de-aliasing'', is to assume
that the illuminant is as close to the center of the histogram as possible.
This de-aliasing approach simply requires carefully setting the starting point
of the histogram $(u_{\mathit{lo}}, v_{\mathit{lo}})$ such that the true
illuminants in natural scenes all lie within the span of the histogram, and
setting $\hat{L}' = \hat{L}$.
We do this by setting $u_{\mathit{lo}}$ and $v_{\mathit{lo}}$
to maximize the distance between the edges
of the histogram and the bounding box surrounding the ground-truth illuminants
in the training data\footnote{Our histograms are shifted toward green colors
rather than centered around a neutral color, as cameras are traditionally
designed with an more sensitive green channel which enables white balance to be
performed by gaining red and blue up without causing color clipping. Ignoring
this practical issue, our approach can be thought of as centering our histograms
around a neutral white light}.
Gray light de-aliasing is trivial to implement but,
 unlike gray world de-aliasing, it will systematically
fail if the histogram is too small to fit all illuminants within its span.

To summarize the difference between CCC \cite{BarronICCV2015} and our approach
with regards to illuminant aliasing, CCC (approximately) performs illuminant
estimation as follows:
\begin{equation}
\begin{bmatrix} \hat{L}_u \\ \hat{L}_v \end{bmatrix} = \begin{bmatrix}u_\mathit{lo} \\ v_\mathit{lo}\end{bmatrix} + \binwidth \left( \argmax_{i, j}\left( N * F \right) \right)
\end{equation}
Where $N * F$ is performed using a pyramid convolution.
FFCC corresponds to this procedure:
\begin{align}
P &\leftarrow \operatorname{softmax} \left( N * F \right) \label{eq:softmax1} \\
(\boldsymbol{\mu}, \boldsymbol{\Sigma}) &\leftarrow \operatorname{fit\_bvm}(P)  \label{eq:bvm1} \\
\begin{bmatrix} \hat{L}_u \\ \hat{L}_v \end{bmatrix} &\leftarrow \operatorname{de\_alias}(\boldsymbol{\mu}) \label{eq:dealias}
\end{align}
Where $N$ is a small and aliased toroidal histogram, convolution is performed with
FFTs, and the centroid of the filtered histogram is estimated and
de-aliased as necessary.
By constructing this pipeline to be differentiable we can train our model in an end-to-end fashion by propagating the gradients of some loss computed on the
de-aliased illuminant prediction $\hat{L}$ back onto the learned filters $F$.
The centroid fitting in Eq.~\ref{eq:bvm1} is performed by fitting
a bivariate von Mises distribution to a PDF, which we will now explain.

\section{Differentiable Bivariate von Mises}
\label{sec:bvm}

\begin{figure}[!]
\centering
  \includegraphics[width=1.05in]{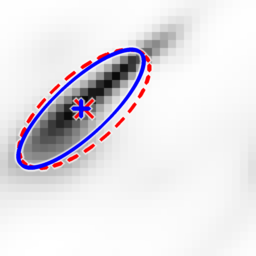}
  \includegraphics[width=1.05in]{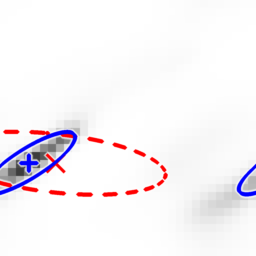}
  \includegraphics[width=1.05in]{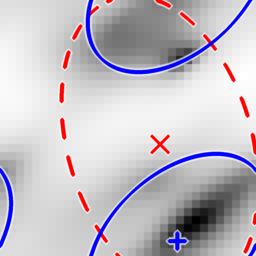}
  \caption{
  We fit a bivariate von Mises distribution (shown in
  solid blue) to toroidal PDFs $P(i,j)$ to produce an aliased illuminant estimate.
  Contrast this with fitting a bivariate Gaussian (shown in dashed red)
  which treats the PDF as if it lies on a plane.
  Both approaches behave similarly if the distribution lies near the center of the
  unwrapped plane (left) but fitting a Gaussian fails as the distribution begins
  to ``wrap around'' the edge (middle, right).
  \label{fig:vonmises}
  }
\end{figure}

Our architecture requires some mechanism for reducing a
toroidal PDF $P(i,j)$ to a single estimate of the illuminant.
Localizing the center of mass of a histogram defined on a torus is difficult:
fitting a bivariate Gaussian may
fail when the input distribution ``wraps around'' the sides of the PDF, as
shown in Fig.~\ref{fig:vonmises}.
Additionally, for the sake of temporal smoothing (Section~\ref{sec:temporal})
and confidence estimation,
we want our model to predict a well-calibrated covariance matrix around the
center of mass of $P$.
This requires that our model be trained end-to-end,
which therefore requires that our mean/covariance fitting be analytically
differentiable and therefore usable as a ``layer'' in our learning
architecture.
To address these problems we present a variant of the bivariate von Mises
distribution~\cite{Mardia1975}, which we will use to efficiently localize
the mean and covariance of $P$ in a manner that allows for easy backpropagation.

The bivariate von Mises distribution (BVM) is a common parameterization of a
PDF on a torus.
There exist several parametrizations
which mostly differ in how ``concentration'' is represented
(``concentration'' having a similar meaning to covariance).
All of these parametrizations present problems in our use case:
none have closed form expressions for maximum likelihood estimators \cite{Hamelryck2012},
none lend themselves to convenient backpropagation,
and all define concentration in terms of angles and therefore
require ``conversion'' to covariance matrices during color de-aliasing.
For these reasons we present an alternative parametrization in which we directly
estimate a BVM as a mean $\boldsymbol{\mu}$ and covariance $\boldsymbol{\Sigma}$
in a simple and differentiable closed form expression.
Though necessarily approximate,
our estimator is accurate when the distribution is well-concentrated,
which is generally the case for our task.

Our input is a PDF $P(i,j)$ of size $n \times n$,
where $i$ and $j$ are integers in $[0, n-1]$.
For convenience we define a mapping from $i$ or $j$ to angles in $[ 0, 2\pi )$
and the marginal distributions of $P$ with respect to $i$ and $j$:
\begin{equation}
\theta(i) = {2\pi i \over n} \quad\,\, P_i(i) = \sum_j P(i, j) \quad\,\, P_j(j) = \sum_i P(i, j) \nonumber
\end{equation}
We also define the marginal expectation of the sine and cosine of the angle:
\begin{equation}
y_i = \sum_i P_i(i) \sin(\theta(i)) \quad x_i = \sum_i P_i(i) \cos(\theta(i))
\end{equation}
With $x_j$ and $y_j$ defined similarly.

Estimating the mean $\boldsymbol \mu$ of a BVM from a histogram just requires
computing the circular mean in $i$ and $j$:
\begin{equation}
\boldsymbol{\mu} = \begin{bmatrix}u_\mathit{lo} \\ v_\mathit{lo}\end{bmatrix} + \binwidth\begin{bmatrix}
\operatorname{mod}\left( {n \over 2\pi} \atantwo (y_i, x_i), n \right) \\
\operatorname{mod}\left( {n \over 2\pi} \atantwo (y_j, x_j), n \right) \\
\end{bmatrix} \label{eq:mu}
\end{equation}
Eq.~\ref{eq:mu} includes gray light de-aliasing, though gray world de-aliasing can also
be applied to $\boldsymbol{\mu}$ after fitting.

We can fit the covariance of our model
by simply ``unwrapping'' the coordinates of the histogram relative to the
estimated mean and treating these unwrapped coordinates
as though we are fitting a bivariate Gaussian.
We define the ``unwrapped'' $(i, j)$ coordinates such that the ``wrap around''
point on the torus lies as far away from the mean as possible,
or equivalently,
such that the unwrapped coordinates are as close to the mean as possible:
\begin{align}
\ishift & = \operatorname{mod}\left( i - \floor{\boldsymbol{\mu}_u - u_\mathit{lo} \over h} + {n \over 2}, n \right) \nonumber \\
\jshift & = \operatorname{mod}\left( j - \floor{\boldsymbol{\mu}_v - v_\mathit{lo} \over h} + {n \over 2}, n \right)
\end{align}
Our estimated covariance matrix is simply the sample covariance of $P(\ishift, \jshift)$:
\begin{equation}
\displaystyle \E{\ishift} = \sum_i P_i(i) \ishift \quad \quad \displaystyle \E{\jshift} = \sum_j P_j(j) \jshift
\end{equation}
\begin{equation}
\resizebox{2.9in}{!}{$
\boldsymbol{\Sigma} = h^2
\begin{bmatrix}
\displaystyle \epsilon + \sum_i P_i(i) \ishift^2 - \E{\ishift}^2&
\displaystyle \sum_{i,j} P(i,j) \ishift \jshift - \E{\ishift} \E{\jshift} \\
\displaystyle \sum_{i,j} P(i,j) \ishift \jshift - \E{\ishift} \E{\jshift} &
\displaystyle \epsilon + \sum_j P_j(j) \jshift^2 - \E{\jshift}^2
\end{bmatrix}
$} \label{eq:covariance}
\end{equation}
We regularize the sample covariance matrix slightly by adding a constant $\epsilon=1$
to the diagonal.

With our estimated mean and covariance we can compute our loss: the negative log-likelihood
of a Gaussian (ignoring scale factors and constants) relative to the true illuminant $L^*$:
\begin{equation}
\resizebox{2.9in}{!}{$
\lossdata{\boldsymbol{\mu}, \boldsymbol{\Sigma}} = \log{| \boldsymbol{\Sigma} | } + \left( \begin{bmatrix}
L^*_u \\ L^*_v
\end{bmatrix} - \boldsymbol{\mu} \right)^\mathrm{T} \boldsymbol{\Sigma}^{-1} \left( \begin{bmatrix}
L^*_u \\ L^*_v
\end{bmatrix} - \boldsymbol{\mu} \right) \label{eq:lossdata}
$}
\end{equation}
Using this loss causes our model to produce a well-calibrated complete posterior
of the illuminant instead of just a single estimate.
This posterior will be useful when processing video sequences
(Section~\ref{sec:temporal})
and also allows us to attach confidence estimates to our predictions
using the entropy of $\boldsymbol{\Sigma}$ (see the appendix).

Our entire system is trained end-to-end, which requires
that every step in BVM fitting and loss computation be
analytically differentiable. See the appendix for the analytical gradients
for Eqs.~\ref{eq:mu}, \ref{eq:covariance}, and \ref{eq:lossdata}, which
can be chained together to backpropagate the gradient of $\lossdata{\cdot}$
onto the input PDF $P$.

\section{Model Extensions}
\label{sec:extensions}

The system we have described thus far (compute a periodic histogram of
each pixel's log-chroma, apply a learned FFT convolution, apply a softmax,
fit a de-aliased bivariate von Mises distribution) works reasonably well
(Model A in Table~\ref{table:gehler_shi})
but does not produce state-of-the-art results.
This is likely because this model reasons about pixels independently,
ignores all spatial information in the image,
and does not consider the absolute color of the illuminant.
Here we present extensions to the model which address these issues
and improve accuracy accordingly.

As explored in \cite{BarronICCV2015}, a CCC-like model
can be generalized to a set of ``augmented'' images provided that these images
are non-negative and ``scale with intensity'' \cite{Finlayson2013}.
This lets us apply certain filtering operations to image $I$ and, instead of
constructing a single histogram from our image,
construct a ``stack'' of histograms constructed from the image and its
filtered versions. Instead of learning
and applying one filter, we learn a stack of filters and sum across
channels after convolution.
The general family of augmented images used in \cite{BarronICCV2015}
are expensive to compute, so we instead use just
the input image $I$ and a local measure of absolute deviation in the input image:
\begin{equation}
\resizebox{2.9in}{!}{$
E(x,y,c) = {1 \over 8} \displaystyle \sum_{i = -1}^1 \sum_{j = -1}^1 \abs{I(x,y,c) - I(x+i,y+j,c)}
$}
\end{equation}
These two features appears to perform similarly to the four features used in
\cite{BarronICCV2015}, while being cheaper to compute.

Just as a sliding-window object detector is often invariant to the absolute location of an
object in an image,
the convolutional nature of our baseline model makes it invariant to any global shift of the color of the input image.
This means that our baseline model \emph{cannot} rely on any statistical
regularities of the illumination by, say,
modeling black body radiation,
the specific properties of commonly manufactured light bulbs,
or any varying spectral sensitivity across cameras.
Though CCC does not model illumination directly,
it appears to indirectly reason about illumination by using the boundary
conditions of its pyramid convolution to learn a model which is not truly
spatially varying and is therefore sensitive to absolute color.
Because a torus has no boundaries, our model is invariant to
global input color, so we must therefore
introduce a mechanism for directly reasoning about illuminants.
We use a per-illuminant ``gain'' map $G(i,j)$ and ``bias'' map $B(i,j)$,
which together apply a per-illuminant affine transformation to the output of our
previously-described convolution at (aliased) color $(i,j)$.
The bias $B$ causes our model
to prefer certain illuminants over others, while the gain $G$ causes the
contribution of the convolution at certain colors to be amplified.

\begin{figure}[!]
\centering
  \stackunder[5pt]
    {\includegraphics[width=0.78in]{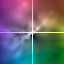}}
    {\footnotesize (a) Pixel Filter \label{fig:filter_pixel}}
  \stackunder[5pt]
    {\includegraphics[width=0.78in]{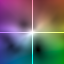}}
    {\footnotesize (b) Edge Filter \label{fig:filter_edge}}
  \stackunder[5pt]
    {\includegraphics[width=0.78in]{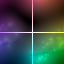}}
    {\footnotesize (c) Illum. Gain \label{fig:blackbody_gain}}
  \stackunder[5pt]
    {\includegraphics[width=0.78in]{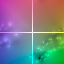}}
    {\footnotesize (d) Illum. Bias \label{fig:blackbody_bias}}
  \caption{
  A complete learned model (Model J in Table~\ref{table:gehler_shi})
  shown in centered $(u,v)$ log-chroma space, with brightness indicating
  larger values.
  Our learned filters are centered around the origin
  (the predicted white point)
  and our illuminant gain and bias maps model the black body curve and
  varying camera sensitivity as two wrap-around line segments (this dataset
  consists of images from two different cameras).
  \label{fig:blackbody}
  }
\end{figure}

Our two extensions (an augmented edge channel and an illuminant gain/bias map)
let us redefine the $P$ in Eq.~\ref{eq:softmax1} as
\begin{equation}
P = \operatorname{softmax} \left(B + G \circ \sum_k \left( N_k * F_k \right) \right)
\end{equation}
Where $\{F_k\}$ are the set of learned filters for each augmented channel's
histogram $N_k$,
$G$ is our learned gain map, and $B$ is our learned bias map.
In practice we actually parametrize $G_{\log}$ when training and define
$G = \exp(G_{\log})$, which constraints $G$ to be non-negative.
Visualizations of $G$ and $B$ and our learned filters can be seen in
Fig.~\ref{fig:blackbody}.

\section{Fourier Regularization and Preconditioning}
\label{sec:fourier}

Our learned model weights $( \{ F_k \}, G, B )$ are all periodic $n \times n$ images.
To improve generalization, we want these weights to be small and smooth.
In this section we present the general form of the regularization used during
training, and we show how this regularization lets us precondition the
optimization problem solved during training to find lower-cost minima in fewer
iterations.
Because this frequency-domain optimization technique applies
generally to any optimization problem concerning smooth and periodic images,
we will describe it in general terms.

Let us construct an optimization problem with respect to a single $n \times n$
image $Z$ consisting of a data term $f(Z)$ and a regularization term $g(Z)$:
\begin{equation}
Z^* = \argmin_Z \left( f\left(Z\right) + g\left(Z\right) \right)
\label{eq:primal_loss}
\end{equation}
We require that the regularization $g(Z)$
is the weighted sum of squared periodic convolutions of $Z$ with some filter bank.
In our experiments $g(Z)$ is the weighted
sum of the squared difference between adjacent values (similar to a total
variation loss \cite{Rudin1992}) and the sum of squared values:
\begin{align}
g(Z) = \textstyle & \lambda_1 \textstyle \sum_{i, j} \big( \left(Z \left(i,j \right) - Z \left(\operatorname{mod}(i+1, n),j \right) \right)^2 \nonumber \\
 & \quad\quad\,\,\,\, + \left( Z\left(i,j\right) - Z\left(i,\operatorname{mod}(j+1, n)\right) \right)^2 \big) \nonumber \\
 + & \lambda_0 \textstyle \sum_{i, j} Z(i,j)^2 \label{eq:reg}
\end{align}
Where $\lambda_1$ and $\lambda_0$ are hyperparameters that determine the
strength of each smoothness term. We require that $\lambda_0 > 0$ to prevent
divide-by-zero issues during preconditioning.

We use a variant of the standard FFT
$\fftv{\cdot}$ which bijectively maps from some real $n \times n$ image to a
real $n^2$-dimensional vector, instead of the complex $n \times n$ image
produced by a standard FFT (See the appendix for a formal description).
With this, we can rewrite Eq.~\ref{eq:reg} as follows:
\begin{align}
&\mathbf{w} = {1 \over n} \sqrt{ \lambda_1 \left( \abs{ \fftv{[1, -1]}}^2 + \abs{\fftv{[1; -1]}}^2 \right) + \lambda_0} \nonumber \\
&g(Z) = \fftv{Z}^\mathrm{T} \diag \left( \mathbf{w} \right)^2  \fftv{Z}
\end{align}
where the vector $\mathbf{w}$ is just some fixed function of the definition
of $g(Z)$ and the values of the hyperparameters $\lambda_1$ and $\lambda_0$.
The 2-tap difference filters in $\fftv{[1, -1]}$ and $\fftv{[1; -1]}$
are padded to size $(n \times n)$ before the FFT.
With $\mathbf{w}$ we can define a mapping between our 2D image space and
a rescaled FFT vector space:
\begin{equation}
\mathbf{z} = \mathbf{w} \circ \fftv{Z}
\end{equation}
Where $\circ$ is an element-wise product.
This mapping lets us rewrite the optimization problem in Eq.~\ref{eq:primal_loss} as:
\begin{equation}
\resizebox{2.9in}{!}{$
Z^* = \ifftv{ {1 \over \mathbf{w}} \left( \displaystyle \argmin_{\mathbf{z}} \left( f \left( \ifftv{ {\mathbf{z} \over \mathbf{w}}} \right) + \norm{\mathbf{z}}^2 \right)} \right)
\label{eq:fourier_loss}
$}
\end{equation}
where $\ifftv{\cdot}$ is the inverse of $\fftv{\cdot}$, and division is element-wise.
This reparametrization reduces the complicated regularization of $Z$
to a simple L2 regularization of $\mathbf{z}$, which has a preconditioning
effect.

\begin{figure}[t!]
\centering
  \stackunder[5pt]
    {\footnotesize Logistic Loss}
    {\includegraphics[width=1.6in]{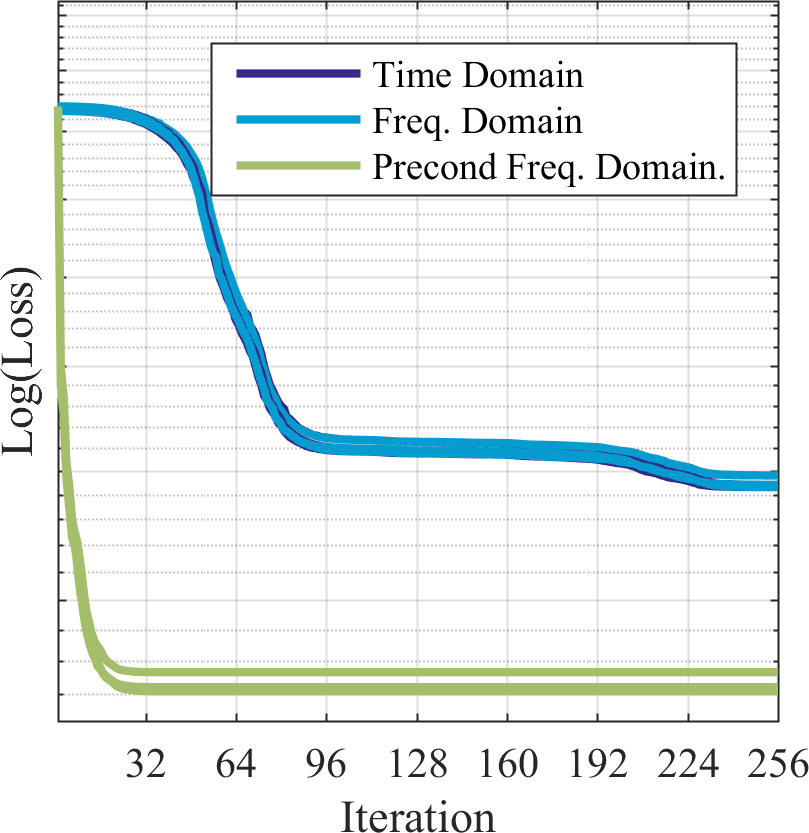}}
  \stackunder[5pt]
    {\footnotesize BVM Loss}
    {\includegraphics[width=1.6in]{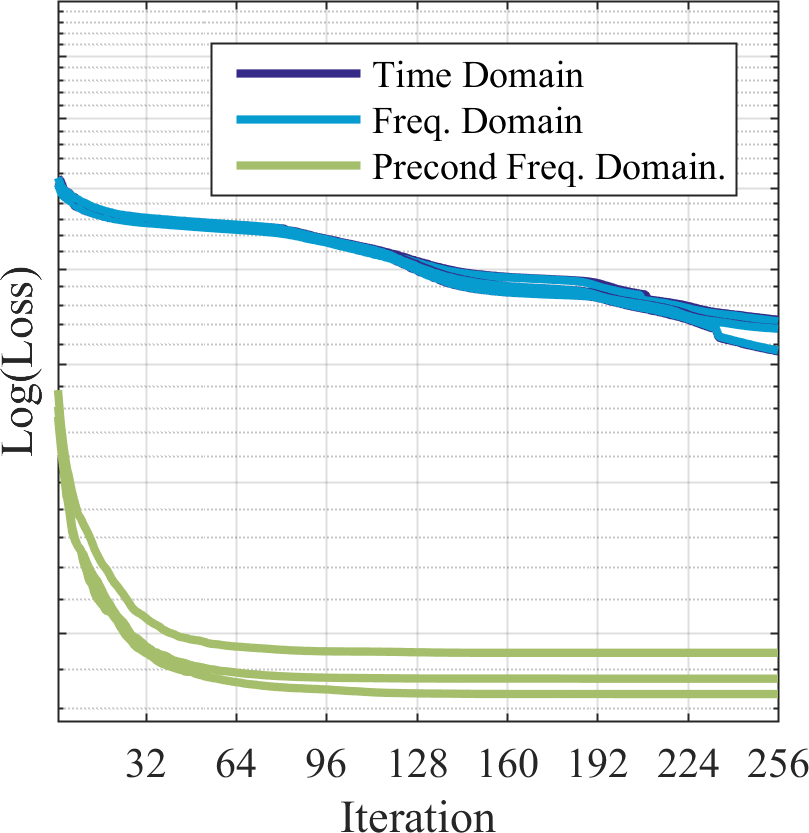}}
  \caption{
  Loss traces for our two stages of training, for three fold
  cross validation (each line represents a fold)
  on the Gehler-Shi dataset using LBFGS. Our preconditioned
  frequency domain optimization produces lower minima at greater rates
  than are achieved by non-preconditioned optimization in the
  frequency domain or naive optimization in the time domain.
  \label{fig:opt}
  }
\end{figure}

We use this technique during training to reparameterize all model
components $( \{F_k \}, G, B )$ as rescaled FFT vectors, each with their own
values for $\lambda_0$ and $\lambda_1$. The effect of this can be seen in
Fig.~\ref{fig:opt}, where we show the loss during our two training stages.
We compare against naive time-domain optimization (Eq.~\ref{eq:primal_loss})
and non-preconditioned frequency-domain optimization (Eq.~\ref{eq:fourier_loss}
with $\mathbf{w}=1$).
Our preconditioned reformulation exhibits a significant speedup and finds
minima with lower losses.

For all experiments (excluding our ``deep'' variants, see the appendix),
training is as follows:
All model parameters are initialized to $0$,
then we have a convex pre-training step which optimizes
Eq.~\ref{eq:fourier_loss} where $f(\cdot)$ is a logistic loss
(described in the appendix) using LBFGS for $16$ iterations,
and then we optimize Eq.~\ref{eq:fourier_loss}
where $f(\cdot)$ is the non-convex BVM loss in Eq.~\ref{eq:lossdata} using
LBFGS for $64$ iterations.

\section{Temporal Smoothing}
\label{sec:temporal}

Color constancy is usually studied in the context of
individual images, which are assumed to be IID.
But a practical white balance algorithm must run on a video sequence,
and must enforce some temporal smoothing of the predicted illuminant
to avoid presenting the viewer with an erratically-varying image in the viewfinder.
This smoothing cannot be too aggressive or else the viewfinder may appear
unresponsive when the illumination changes rapidly
(a colorful light turning on, the camera quickly moving outdoors, etc).
Additionally, when faced with multiple valid hypotheses
(a blue wall under white light vs a white wall under blue light, etc) we
may want to use earlier images to resolve ambiguities.
These desiderata of stability, responsiveness, and robustness are at odds
with each other, and so some compromise must be struck.

Our task of constructing a temporally coherent illuminant estimate is aided
by the probabilistic nature of the output of our per-frame model, which
produces a posterior distribution over illuminants parametrized as a bivariate
Gaussian.
Let us assume that we have some ongoing estimate of the illuminant and its
covariance $(\boldsymbol{\mu}_t, \boldsymbol{\Sigma}_t)$.
Given the observed mean and covariance $(\boldsymbol{\mu}_o, \boldsymbol{\Sigma}_o)$
provided by our model we update our
ongoing estimate by first convolving it with an zero-mean isotropic Gaussian
(encoding our prior belief that the illuminant may change over time)
and then multiplying that ``fuzzed'' Gaussian by the observed Gaussian:
\begin{align}
\boldsymbol{\Sigma}_{t+1} &= \left( \left( \boldsymbol{\Sigma}_t + \begin{bmatrix}\alpha&0\\0&\alpha\end{bmatrix} \right)^{-1} + \boldsymbol{\Sigma}_o \right)^{-1} \\
\boldsymbol{\mu}_{t+1} &= \boldsymbol{\Sigma}_{t+1} \left( \left( \boldsymbol{\Sigma}_t + \begin{bmatrix}\alpha&0\\0&\alpha\end{bmatrix} \right)^{-1} \boldsymbol{\mu}_t + \boldsymbol{\Sigma}_o \boldsymbol{\mu}_o \right)^{-1} \nonumber
\end{align}
Where $\alpha$ is a parameter that defines the expected variance of the
illuminant over time.
This update resembles a Kalman filter but with a simplified transition model,
no control model, and variable observation noise.

This temporal smoothing is not used in our benchmarks, but
its effect can be seen in the supplemental video.

\section{Results}
\label{sec:results}

We evaluate our technique using two standard color constancy datasets:
the Gehler-Shi dataset \cite{Gehler08,shifunt}
and the Cheng \etal\ dataset \cite{Cheng14} (see Tables~\ref{table:gehler_shi}
and~\ref{table:cheng}).
For the Gehler-Shi dataset we present several ablations and variants
of our model to show the effect of each design decision
and to investigate trade-offs between speed and accuracy.
Models labeled ``full'' were run on $384 \times 256$ $16$-bit images,
while models labeled ``thumb'' were run on $48 \times 32$ $8$-bit images,
which are the kind of images that
a practical white-balance system embedded on a hardware device might use.
Models labeled ``4 channel'' use the four feature channels used in
\cite{BarronICCV2015}, while models labeled ``2 channel'' use the two
channels we present in Section~\ref{sec:extensions}.
We also present models in which we only use the ``pixel channel'' $I$
or the ``edge channel'' $E$ as input.
All models have a histogram size of $n=64$ except for Models K and L where
$n$ is varied to show the impact of illuminant aliasing.
Two models use ``gray world'' de-aliasing, and the rest use ``gray light''
de-aliasing.
The former seems slightly less effective than the latter unless chroma histograms are heavily aliased, which is why we use it in Model K.
Model C only has one training stage that minimizes logistic loss for $64$ iterations,
thereby removing the BVM fitting from training.
Model E fixes $G(i,j) = 1$ and $B(i,j) = 0$, thereby removing
the model's ability to reason about the absolute color of the illuminant.
Model B was trained only to minimize the data term
(ie, $\lambda_0 = \lambda_1 = 0$ in Eq.~\ref{eq:reg}) while
Model D uses L2 regularization but not total variation
(ie, $\lambda_1 = 0$ in Eq.~\ref{eq:reg}).
Models N, O and P are variants of Model J in which, instead of learning a fixed model $( \{ F_k \}, G, B )$
we express those model parameters as the output of a small 2-layer neural network.
As inputs to this network we use image metadata, which allows the model to reason about exposure time and camera sensor type,
and/or a CNN-produced feature vector \cite{Wang2014}, which allows the model to reason about semantics (see the appendix for details).
For each experiment we tune all $\lambda$ hyperparameters to minimize the
``average'' error during cross-validation, using cyclic coordinate descent.

Model P achieves the lowest-error results,
with a $20\%$ reduction in error on Gehler-Shi compared to the previously best-performing published technique.
This improvement in accuracy also comes with a significant speedup compared to
previous techniques: $\sim\!30$ ms/image for most models, compared to the $520$ ms of CCC
\cite{BarronICCV2015} or the $3$ seconds (on a GPU) of Shi \etal\ \cite{ShiECCV2016}.
Model Q (our fastest model) has an accuracy comparable to
\cite{BarronICCV2015} and \cite{ShiECCV2016} but takes only $1.1$ milliseconds
to process an image, making it hundreds or millions of times faster than the
current state-of-the art.
Additionally, our model appears to be faster to train than the state-of-the-art,
though training times for prior work are often not available.
All runtimes in Table~\ref{table:gehler_shi} for our model were computed on an
Intel Xeon CPU E5-2680.
Runtimes for the ``full'' model were produced using a
Matlab implementation, while runtimes for the ``thumb'' model were produced
using a Halide \cite{RaganKelley2012} CPU implementation
(our Matlab implementation of Model Q takes $2.37$ ms/image).
Runtimes for our ``+semantic'' models are not presented as we were unable to
profile \cite{Wang2014} accurately (CNN feature computation appears to dominate
runtime).

To demonstrate that our model is a viable automatic white
balance system for consumer photography, we ran our Halide code on a
2016 Google Pixel XL using the thumbnail images
computed by the device's camera stack.
This implementation ran at $1.44$ms per image, which is equivalent
to $30$ frames per second using $< 5\%$ of the total compute
budget, thereby satisfying our previously-stated speed requirements.
A video of our system running in real-time on a phone can be found in the
supplement.

\begin{table}[!]
\begin{center}
\resizebox{3.25in}{!}{
\Huge
\begin{tabular}{ l  | c c c c c | c || c c}
\multirow{2}{*}{Algorithm} & \multirow{2}{*}{Mean} & \multirow{2}{*}{Med.} & \multirow{2}{*}{Tri.} & Best & Worst & \multirow{2}{*}{Avg.} & Test & Train \\
 & & & & 25\% & 25\% & & Time & Time \\
\hline
Support Vector Regression \cite{FuntX04} & $ 8.08 $ & $ 6.73 $ & $ 7.19 $ & $ 3.35 $ & $ 14.89 $ & $ 7.21 $ &  - &  - \\
White-Patch \cite{Brainard86} & $ 7.55 $ & $ 5.68 $ & $ 6.35 $ & $ 1.45 $ & $ 16.12 $ & $ 5.76 $ & $ 0.16 $ &  - \\
Grey-world \cite{Buchsbaum80} & $ 6.36 $ & $ 6.28 $ & $ 6.28 $ & $ 2.33 $ & $ 10.58 $ & $ 5.73 $ & $ 0.15 $ &  - \\
Edge-based Gamut \cite{Gijsenij2010} & $ 6.52 $ & $ 5.04 $ & $ 5.43 $ & $ 1.90 $ & $ 13.58 $ & $ 5.40 $ & $ 3.6 $ & $ 1986 $\\
1st-order Gray-Edge \cite{vandeWeijerTIP2007} & $ 5.33 $ & $ 4.52 $ & $ 4.73 $ & $ 1.86 $ & $ 10.03 $ & $ 4.63 $ & $ 1.1 $ &  - \\
2nd-order Gray-Edge \cite{vandeWeijerTIP2007} & $ 5.13 $ & $ 4.44 $ & $ 4.62 $ & $ 2.11 $ & $ 9.26 $ & $ 4.60 $ & $ 1.3 $ &  - \\
Shades-of-Gray \cite{FinlaysonT04} & $ 4.93 $ & $ 4.01 $ & $ 4.23 $ & $ 1.14 $ & $ 10.20 $ & $ 3.96 $ & $ 0.47 $ &  - \\
Bayesian \cite{Gehler08} & $ 4.82 $ & $ 3.46 $ & $ 3.88 $ & $ 1.26 $ & $ 10.49 $ & $ 3.86 $ & $ 97 $ & $ 764 $\\
Yang \etal\, 2015 \cite{Yang2015} & $ 4.60 $ & $ 3.10 $ & - & - & - & - & $ 0.88 $ &  - \\
General Gray-World \cite{Barnard02} & $ 4.66 $ & $ 3.48 $ & $ 3.81 $ & $ 1.00 $ & $ 10.09 $ & $ 3.62 $ & $ 0.91 $ &  - \\
Natural Image Statistics \cite{GijsenijTPAMI2011} & $ 4.19 $ & $ 3.13 $ & $ 3.45 $ & $ 1.00 $ & $ 9.22 $ & $ 3.34 $ & $ 1.5 $ & $ 10749 $\\
CART-based Combination \cite{Bianco2010} & $ 3.90 $ & $ 2.91 $ & $ 3.21 $ & $ 1.02 $ & $ 8.27 $ & $ 3.14 $ &  - &  - \\
Spatio-spectral Statistics \cite{Chakrabarti11} & $ 3.59 $ & $ 2.96 $ & $ 3.10 $ & $ 0.95 $ & $ 7.61 $ & $ 2.99 $ & $ 6.9 $ & $ 3159 $\\
LSRS \cite{Gao2014} & $ 3.31 $ & $ 2.80 $ & $ 2.87 $ & $ 1.14 $ & $ 6.39 $ & $ 2.87 $ & $ 2.6 $ & $ 1345 $\\
Interesection-based Gamut \cite{Gijsenij2010} & $ 4.20 $ & $ 2.39 $ & $ 2.93 $ & $ 0.51 $ & $ 10.70 $ & $ 2.76 $ &  - &  - \\
Pixels-based Gamut \cite{Gijsenij2010} & $ 4.20 $ & $ 2.33 $ & $ 2.91 $ & $ 0.50 $ & $ 10.72 $ & $ 2.73 $ &  - &  - \\
Bottom-up+Top-down \cite{VSV2007a} & $ 3.48 $ & $ 2.47 $ & $ 2.61 $ & $ 0.84 $ & $ 8.01 $ & $ 2.73 $ &  - &  - \\
Cheng \etal\, 2014 \cite{Cheng14} & $ 3.52 $ & $ 2.14 $ & $ 2.47 $ & $ 0.50 $ & $ 8.74 $ & $ 2.41 $ & $ 0.24 $ &  - \\
Exemplar-based \cite{Joze2014} & $ 2.89 $ & $ 2.27 $ & $ 2.42 $ & $ 0.82 $ & $ 5.97 $ & $ 2.39 $ &  - &  - \\
Bianco \etal\, 2015 \cite{Bianco2015} & $ 2.63 $ & $ 1.98 $ & - & - & - & - &  - &  - \\
Corrected-Moment \cite{Finlayson2013} & $ 2.86 $ & $ 2.04 $ & $ 2.22 $ & $ 0.70 $ & $ 6.34 $ & $ 2.25 $ & $ 0.77 $ & $ 584 $\\
Chakrabarti \etal\, 2015 \cite{Chakrabarti2015} & $ 2.56 $ & $ 1.67 $ & $ 1.89 $ & $ 0.52 $ & $ 6.07 $ & $ 1.91 $ & $ 0.30 $ &  - \\
Cheng \etal\, 2015 \cite{ChengCVPR2015} & $ 2.42 $ & $ 1.65 $ & $ 1.75 $ & $ 0.38 $ & $ 5.87 $ & $ 1.73 $ & $ 0.25 $ & $ 245 $\\
CCC \cite{BarronICCV2015} & $ 1.95 $ & $ 1.22 $ & $ 1.38 $ & $ 0.35 $ & $ 4.76 $ & $ 1.40 $ & $ 0.52 $ & $ 2168 $\\
Shi \etal\, 2016 \cite{ShiECCV2016} & $ 1.90 $ & $ 1.12 $ & $ 1.33 $ & $ 0.31 $ & $ 4.84 $ & $ 1.34 $ & $ 3.0 $ &  - \\
\hline
\texttt{A}) FFCC - full, pixel channel only, no illum. & $ 2.88 $ & $ 1.90 $ & $ 2.05 $ & $ 0.50 $ & $ 6.98 $ & $ 2.08 $ & $ 0.0076 $ & $ 117 $\\
\texttt{B}) FFCC - full 2 channels, no regularization & $ 2.34 $ & $ 1.33 $ & $ 1.55 $ & $ 0.51 $ & $ 5.84 $ & $ 1.70 $ & $ 0.031 $ & $ 96 $\\
\texttt{C}) FFCC - full 2 channels, no BVM loss & $ 2.16 $ & $ 1.45 $ & $ 1.56 $ & $ 0.76 $ & $ 4.84 $ & $ 1.78 $ & $ 0.031 $ & $ 62 $\\
\texttt{D}) FFCC - full 2 channels, no total variation & $ 1.92 $ & $ 1.11 $ & $ 1.27 $ & $ 0.28 $ & $ 4.89 $ & $ 1.30 $ & $ 0.028 $ & $ 104 $\\
\texttt{E}) FFCC - full, 2 channels, no illuminant & $ 2.14 $ & $ 1.34 $ & $ 1.52 $ & $ 0.37 $ & $ 5.27 $ & $ 1.53 $ & $ 0.031 $ & $ 94 $\\
\texttt{F}) FFCC - full, pixel channel only & $ 2.15 $ & $ 1.33 $ & $ 1.51 $ & $ 0.34 $ & $ 5.35 $ & $ 1.51 $ & $ 0.0063 $ & $ 67 $\\
\texttt{G}) FFCC - full, edge channel only & $ 2.02 $ & $ 1.25 $ & $ 1.39 $ & $ 0.34 $ & $ 5.11 $ & $ 1.44 $ & $ 0.026 $ & $ 94 $\\
\texttt{H}) FFCC - full, 2 channels, no precond. & $ 2.91 $ & $ 1.99 $ & $ 2.23 $ & $ 0.57 $ & $ 6.74 $ & $ 2.18 $ & $ 0.025 $ & $ 152 $\\
\texttt{I}) FFCC - full, 2 channels, gray world & $ 1.79 $ & $ 1.01 $ & $ 1.22 $ & $ 0.29 $ & $ 4.54 $ & $ 1.24 $ & $ 0.029 $ & $ 98 $\\
\texttt{J}) FFCC - full, 2 channels & $ 1.80 $ & $ 0.95 $ & $ 1.18 $ & $ 0.27 $ & $ 4.65 $ & $ 1.20 $ & $ 0.029 $ & $ 98 $\\
\texttt{K}) FFCC - full, 4 channels, $n=32$, gray world & $ 2.69 $ & $ 1.31 $ & $ 1.49 $ & $ 0.37 $ & $ 7.48 $ & $ 1.70 $ & $ 0.068 $ & $ 138 $\\
\texttt{L}) FFCC - full, 4 channels, $n=256$ & $ 1.78 $ & $ 1.05 $ & $ 1.19 $ & $ 0.27 $ & $ 4.46 $ & $ 1.22 $ & $ 0.068 $ & $ 395 $\\
\texttt{M}) FFCC - full, 4 channels & $ 1.78 $ & $ 0.96 $ & $ 1.14 $ & $ 0.29 $ & $ 4.62 $ & $ 1.21 $ & $ 0.070 $ & $ 96 $\\
\texttt{N}) FFCC - full, 2 channels, +semantics\cite{Wang2014} & $ 1.67 $ & $ 0.96 $ & $ 1.13 $ & $ 0.26 $ &  \cellcolor{Yellow} $ 4.23 $ & $ 1.15 $ &  - &  - \\
\texttt{O}) FFCC - full, 2 channels, +metadata & $ 1.65 $ & $ 0.86 $ & $ 1.07 $ & $ 0.24 $ & $ 4.44 $ & $ 1.10 $ & $ 0.036 $ & $ 143 $\\
\texttt{P}) FFCC - full, 2 channels, +metadata +semantics\cite{Wang2014} &  \cellcolor{Yellow} $ 1.61 $ &  \cellcolor{Yellow} $ 0.86 $ &  \cellcolor{Yellow} $ 1.02 $ &  \cellcolor{Yellow} $ 0.23 $ & $ 4.27 $ &  \cellcolor{Yellow} $ 1.07 $ &  - &  - \\
\hline
\texttt{Q}) FFCC - thumb, 2 channels & $ 2.01 $ & $ 1.13 $ & $ 1.38 $ & $ 0.30 $ & $ 5.14 $ & $ 1.37 $ & $ 0.0011 $ & $ 73 $

\end{tabular}
}
\vspace{1mm}
\caption{
Performance on the Gehler-Shi dataset \cite{Gehler08,shifunt}.
We present five error metrics and their average
(the geometric mean) with the lowest error per metric
highlighted in yellow.
We present the time (in seconds) for training each model and for
evaluating a single image, when available.
\label{table:gehler_shi}}
\vspace{5mm}
\resizebox{3.25in}{!}{
\Huge
\begin{tabular}{ l  | c c c c c | c}
\multirow{2}{*}{Algorithm} & \multirow{2}{*}{Mean} & \multirow{2}{*}{Med.} & \multirow{2}{*}{Tri.} & Best & Worst & \multirow{2}{*}{Avg.} \\
 & & & & 25\% & 25\% &  \\
\hline
White-Patch \cite{Brainard86} & $ 9.91 $ & $ 7.44 $ & $ 8.78 $ & $ 1.44 $ & $ 21.27 $ & $ 7.24 $ \\
Pixels-based Gamut \cite{Gijsenij2010} & $ 5.27 $ & $ 4.26 $ & $ 4.45 $ & $ 1.28 $ & $ 11.16 $ & $ 4.27 $ \\
Grey-world \cite{Buchsbaum80} & $ 4.59 $ & $ 3.46 $ & $ 3.81 $ & $ 1.16 $ & $ 9.85 $ & $ 3.70 $ \\
Edge-based Gamut \cite{Gijsenij2010} & $ 4.40 $ & $ 3.30 $ & $ 3.45 $ & $ 0.99 $ & $ 9.83 $ & $ 3.45 $ \\
Shades-of-Gray \cite{FinlaysonT04} & $ 3.67 $ & $ 2.94 $ & $ 3.03 $ & $ 0.98 $ & $ 7.75 $ & $ 3.01 $ \\
Natural Image Statistics \cite{GijsenijTPAMI2011} & $ 3.45 $ & $ 2.88 $ & $ 2.95 $ & $ 0.83 $ & $ 7.18 $ & $ 2.81 $ \\
Local Surface Reflectance Statistics \cite{Gao2014} & $ 3.45 $ & $ 2.51 $ & $ 2.70 $ & $ 0.98 $ & $ 7.32 $ & $ 2.79 $ \\
2nd-order Gray-Edge \cite{vandeWeijerTIP2007} & $ 3.36 $ & $ 2.70 $ & $ 2.80 $ & $ 0.89 $ & $ 7.14 $ & $ 2.76 $ \\
1st-order Gray-Edge \cite{vandeWeijerTIP2007} & $ 3.35 $ & $ 2.58 $ & $ 2.76 $ & $ 0.79 $ & $ 7.18 $ & $ 2.67 $ \\
Bayesian \cite{Gehler08} & $ 3.50 $ & $ 2.36 $ & $ 2.57 $ & $ 0.78 $ & $ 8.02 $ & $ 2.66 $ \\
General Gray-World \cite{Barnard02} & $ 3.20 $ & $ 2.56 $ & $ 2.68 $ & $ 0.85 $ & $ 6.68 $ & $ 2.63 $ \\
Spatio-spectral Statistics \cite{Chakrabarti11} & $ 3.06 $ & $ 2.58 $ & $ 2.74 $ & $ 0.87 $ & $ 6.17 $ & $ 2.59 $ \\
Bright-and-dark Colors PCA \cite{Cheng14} & $ 2.93 $ & $ 2.33 $ & $ 2.42 $ & $ 0.78 $ & $ 6.13 $ & $ 2.40 $ \\
Corrected-Moment \cite{Finlayson2013} & $ 2.95 $ & $ 2.05 $ & $ 2.16 $ & $ 0.59 $ & $ 6.89 $ & $ 2.21 $ \\
Color Dog \cite{BanicL15} & $ 2.83 $ & $ 1.77 $ & $ 2.03 $ & $ 0.48 $ & $ 7.04 $ & $ 2.03 $ \\
Shi \etal\, 2016 \cite{ShiECCV2016} & $ 2.24 $ & $ 1.46 $ & $ 1.68 $ & $ 0.48 $ & $ 6.08 $ & $ 1.74 $ \\
CCC \cite{BarronICCV2015} & $ 2.38 $ & $ 1.48 $ & $ 1.69 $ & $ 0.45 $ & $ 5.85 $ & $ 1.74 $ \\
Cheng 2015 \cite{ChengCVPR2015} & $ 2.18 $ & $ 1.48 $ & $ 1.64 $ & $ 0.46 $ & $ 5.03 $ & $ 1.65 $ \\
\hline
\texttt{M}) FFCC - full, 4 channels &  \cellcolor{Yellow} $ 1.99 $ &  \cellcolor{Yellow} $ 1.31 $ &  \cellcolor{Yellow} $ 1.43 $ &  \cellcolor{Yellow} $ 0.35 $ &  \cellcolor{Yellow} $ 4.75 $ &  \cellcolor{Yellow} $ 1.44 $ \\
\hline
\texttt{Q}) FFCC - thumb, 2 channels & $ 2.06 $ & $ 1.39 $ & $ 1.53 $ & $ 0.39 $ & $ 4.80 $ & $ 1.53 $

\end{tabular}
}
\vspace{1mm}
\caption{
 Performance on the dataset from Cheng \etal \cite{Cheng14}, in the same format as Table~\ref{table:gehler_shi}, excluding runtimes. As was done in \cite{BarronICCV2015} we present the
 average performance (the geometric mean) over all $8$ cameras in the dataset.
\label{table:cheng}}
\end{center}
\end{table}

\section{Conclusion}

We have presented FFCC, a color constancy algorithm that produces a $13-20\%$
reduction in error and a $250-3000\times$ speedup relative to prior work.
In doing so we have introduced the concept of convolutional color
constancy on a torus, and we have introduced techniques for illuminant de-aliasing and
differentiable bivariate von Mises fitting required for this toroidal approach.
We have also presented a novel technique for fast Fourier-domain
optimization subject to a certain family of regularizers.
FFCC produces a complete posterior distribution over illuminants,
which lets us assess the model's confidence and also enables a Kalman
filter-like temporal smoothing model.
FFCC's speed, accuracy, and temporal consistency allows it to be
used for real-time white balance on a consumer camera.

\bibliographystyle{ieee}
\bibliography{fccc}

\clearpage

\appendix

\section{Pretraining}
In the paper we described the data term for our loss function $f(\cdot)$ which
takes a toroidal PDF $P(i,j)$, fits a bivariate von Mises
distribution to $P$, and then computes the negative log-likelihood
of the true white point $L^*$ under that distribution.
This loss is non-convex, and therefore may behave erratically
in the earliest training iterations. This issue is compounded by our
differentiable BVM fitting procedure, which may be inacurate when $P$ has
a low concentration, which is often the case in early iterations.
For this reason, we train our model in two stages:
In the ``pretraining'' stage we replace the data term in our loss function
with a more simple loss: straightforward logistic regression with respect to
$P$ and some ground-truth PDF $P^*$ (Eq.~\ref{eq:losspre}), and then
in the second training stage we use the data term described in the paper while
using the output of pretraining to initialize the model.
Because our regularization is also convex, using this pretraining loss makes
our entire optimization problem convex and therefore straightforward to
optimize, and (when coupled with our use of LBFGS for optimization instead
of some SGD-like approach) also makes training deterministic.

Computing a logistic loss is straightforward: we compute a ground-truth PDF
$P^*$ from the ground-truth illuminant $L^*$, and then compute a standard
logistic loss.
\begin{align}
P^*(i,j) =& \operatorname{mod} \left( { L_u^* - u_\mathit{lo} \over \binwidth} - i, n \right) < 1 \\
\wedge& \operatorname{mod} \left( { L_v^* - v_\mathit{lo} \over \binwidth} - j, n \right) < 1 \nonumber \\
f_{\mathrm{pretrain}}(P) &= -\sum_{i,j} P^*(i,j) \log(P(i,j)) \label{eq:losspre}
\end{align}
This loss behaves very similarly to the loss used in CCC \cite{BarronICCV2015},
but it has the added benefit of being convex.

\section{Backpropagation}

The bivariate von Mises estimation procedure described in the paper can be
thought of as a ``layer'' in a deep learning architecture, as our end-to-end
training procedure requires that we be able to backpropagate through the fitting
procedure and the loss computation. Here we present the gradients of the critical
equations described in the paper.

\begin{equation}
\nabla_{\boldsymbol{\mu}} \lossdata{\boldsymbol{\mu}, \boldsymbol{\Sigma}} = -2 \boldsymbol{\Sigma}^{-1} \left( \begin{bmatrix}
L^*_u \\ L^*_v
\end{bmatrix} - \boldsymbol{\mu} \right)
\end{equation}
\begin{equation}
\resizebox{3.25in}{!}{$
\nabla_{\boldsymbol{\Sigma}} \lossdata{\boldsymbol{\mu}, \boldsymbol{\Sigma}}  = \boldsymbol{\Sigma}^{-1} - \boldsymbol{\Sigma}^{-1} \left( \begin{bmatrix}
L^*_u \\ L^*_v
\end{bmatrix} - \boldsymbol{\mu} \right) \left( \begin{bmatrix}
L^*_u \\ L^*_v
\end{bmatrix} - \boldsymbol{\mu} \right)^\mathrm{T} \boldsymbol{\Sigma}^{-1} \nonumber
$}
\end{equation}
\begin{equation}
\nabla_{P(i,j)}\boldsymbol{\mu} =
\left({n \binwidth \over 2\pi}\right) \begin{bmatrix}
{x_i \sin(\theta(i)) - y_i \cos(\theta(i)) \over x_i^2 + y_i^2} \\
{x_j \sin(\theta(j)) - y_j \cos(\theta(j)) \over x_j^2 + y_j^2}
\end{bmatrix}
\end{equation}
\begin{equation}
\resizebox{3.25in}{!}{$
\nabla_{P(i,j)}\boldsymbol{\Sigma} = h^2 \begin{bmatrix}
\ishift \left( \ishift - 2\E{\ishift} \right), \quad
\left( \ishift - \E{\ishift} \right) \left( \jshift - \E{\jshift} \right) - \E{\ishift}\E{\jshift} \\
\left( \ishift - \E{\ishift} \right) \left( \jshift - \E{\jshift} \right) - \E{\ishift}\E{\jshift}, \quad
\jshift \left( \jshift - 2\E{\jshift} \right) \nonumber
\end{bmatrix}
$}
\end{equation}
By chaining these gradients together we can backpropagate the gradient of the loss
back onto the input PDF $P$. Backpropagating through the softmax operation and the convolution
(and illumination gain/bias) is straightforward and so is not detailed here.

\section{Deep Models}

In the main paper we stated that Models N, O, and P use an alternative parametrization
to incorporate external features during training and testing.
This parameterization allows our model to reason about things other than
simple pixel and edge log-chroma histograms, like semantics and camera metadata.
In the basic model presented in the main paper, we learn a set of weights
$( \{F_k \}, G, B )$, where these weights determine the shape of the filters
used during convolution and the per-color gain/bias applied to the output of
that convolution.
Let us abstractly refer to the concatenation of these (preconditioned,
Fourier-domain) weights as $\mathbf{w}$, and let the loss contributed by the data term
for training data instance $i$ be $f_i(\mathbf{w})$
(here $f_i(\mathbf{w})$ does not just apply a loss, but first undoes the
preconditioning transformation and maps from our real FFT vector space to a
complex 2D FFT).
Training our basic model can be thought of as simply finding
\begin{equation}
\argmin_{\mathbf{w}} \sum_i f_i \left( \mathbf{w} \right)
\end{equation}
To generalize our model, instead of learning a single model $\mathbf{w}$, we
instead define a feature vector for each training instance $\mathbf{x}_i$
and learn a mapping from each $\mathbf{x}_i$ to some $\mathbf{w}_i$ such that
the loss for all $\{ \mathbf{w}_i \}$ is minimized.
Instead of learning a single $\mathbf{w}$, we learn the weights in a small
$2$-layer neural network with a ReLU activation function, where those
network weights define the mapping from features to FFCC parameters.
The resulting optimization problem during training is:
\begin{equation}
\argmin_{\mathbf{W}_1, \mathbf{b}_1, \mathbf{W}_2, \mathbf{b}_2} \sum_i f_i \left( \mathbf{W}_2 \max(0, \mathbf{W}_1 \mathbf{x}_i + \mathbf{b}_1) + \mathbf{b}_2 \right)
\end{equation}
Like in all other experiments we train using batch L-BFGS, but instead of the
two-stage training used in the shallow model (a convex ``pretraining''
loss and a nonconvex final loss), we have only one training stage:
$64$ iterations of LBFGS, in which we minimize a weighted sum of the two losses.
Our input vectors $\{ \mathbf{x}_i \}$ are whitened before training, and
the whitening transformation is absorbed into $\mathbf{W}_1$ and $\mathbf{b}_1$
after training so that unwhitened features can be used at test-time.
Our weights are initialized to random Gaussian noise, unlike the shallow model
which is initialized to all zeros.
Unlike our ``shallow'' model, in which $\mathbf{w}$ is regularized during training,
for our ``deep'' models we do not directly regularize each $\mathbf{w}_i$ but instead
indirectly regularize all $\mathbf{w}_i$ by minimizing the squared 2-norm of each $\mathbf{W}_i$ and $\mathbf{b}_i$.
This use of weight decay to regularize our model depends critically on the
frequency-domain preconditioning we use, which causes a simple weight decay to
indirectly impose the careful smoothness regularizer that was constructed for our shallow model.
Note that our ``deep'' model is equivalent to our ``shallow'' model if the input vector is empty
(ie, $\mathbf{x}_i = [\,]$), as $\mathbf{b}_2$ would behave equivalently to $\mathbf{w}$ in that case.
We use $4$ hidden units for Models N and O, and $8$ hidden units for Model P
(which uses the concatenated features from both Models N and O).
The magnitude of the noise used for initialization and of the weight decay for
each layer of the network are tuned using cross-validation.

To produce the ``metadata'' features used in Models O and P we use the EXIF tags
included in the Gehler-Shi dataset.
Using external information in this way is unusual in the color constancy
literature, which is why this aspect of our model is relegated to just two
experiments (all figures and other results do not use external metadata).
In contrast, camera manufacturers spend significant effort considering sensor
spectral properties and other sources of information that may be useful when
building a white balance system.
For example, knowing that two images came from two different sensors (as is the
case in the Gehler-Shi dataset) allows for a more careful treatment of absolute
color and black body radiation.
And knowing the absolute brightness of the scene (indicated by the camera's
exposure time, etc) can be a useful cue for distinguishing between the bright
light of the sun and the relatively low light of man made light sources.
As the improved performance of Model O demonstrates, this other information is
indeed informative and can induce a significant reduction in error.
We use a compact feature vector that encodes the outer product of the
exposure settings of the camera and the name of the camera sensor itself, all
extracted from the EXIF tags included in the public dataset:
\begin{align}
\mathbf{x}_i =&  \mathrm {vec} ( \\
& [\log(\mathrm{shutter\_speed}_i); \log(\mathrm{f\_number}_i); 1] \nonumber \\
\times & [\mathbf {1}_{\mathrm{Canon1D}} (\mathrm{camera}_i), \mathbf {1}_{\mathrm{Canon5D}} (\mathrm{camera}_i), 1]) \nonumber
\end{align}
Note that the Gehler-Shi dataset uses images from two different Canon cameras,
as reflected here.
The log of the shutter speed and F number are chosen as features because,
in theory, their difference should be proportional to the log of the exposure
value of the image, which should indicate the amount of light receiving by the
camera sensor.

The ``semantics'' features used in Models N and P are simply the output of the
CNN model used in \cite{Wang2014}, which was run on the pre-whitebalance image
after it is center-cropped to a square and resized to $256 \times 256$.
Because this image is in the sensor colorspace, before passing it to the CNN
we scale the green channel by $0.6$, apply a CCM, and apply an sRGB gamma
curve.
These semantic features have a modest positive effect.

\section{Real Bijective FFT}

In the paper we describe $\fftv{Z}$, a FFT variant
that takes the 2D FFT of a $n \times n$ real-valued 2D image $Z$ and then
linearizes it into a real-valued vector with no redundant values.
Having this FFT-like one-to-one mapping between real 2D images and real 1D vectors
enables our frequency-domain preconditioner.

Our modified FFT function is defined as:
\begin{align}
\fftv{Z} &= \left[ \begin{array}{l}
\operatorname {Re} (\fft{Z}(0 : \nicefrac{n}{2}, 0)) \\
\operatorname {Re} (\fft{Z}(0 : \nicefrac{n}{2}, \nicefrac{n}{2})) \\
\operatorname {Re} (\fft{Z}(0 : (n-1), 1:(\nicefrac{n}{2}-1))) \\
\operatorname {Im} (\fft{Z}(1 : (\nicefrac{n}{2}-1), 0)) \\
\operatorname {Im} (\fft{Z}(1 : (\nicefrac{n}{2}-1), \nicefrac{n}{2}-1)) \\
\operatorname {Im} (\fft{Z}(0 : (n-1), 1:(\nicefrac{n}{2}-1))) \\
\end{array} \right]
\end{align}
Where $\fft{Z}(i, j)$ is the complex number at the zero-indexed $(i,j)$
position in the FFT of $Z$, and
 $\operatorname {Re}(\cdot)$ and
$\operatorname {Im}(\cdot)$ extract real and imaginary components, respectively.
The output of $\fftv{Z}$ is an $n^2$-dimensional vector, as it must be for our
mapping to preserve all FFT coefficients with no redundancy.
To preserve the scale of the FFT through this mapping we scale $\fftv{Z}$ by
$\sqrt{2}$, ignoring the entries that correspond to:
\begin{align}
&\operatorname {Re}(\fft{Z}(0, 0)) \nonumber \\
&\operatorname {Re}(\fft{Z}(0, \nicefrac{n}{2})) \nonumber \\
&\operatorname {Re}(\fft{Z}(\nicefrac{n}{2}, 0)) \nonumber \\
&\operatorname {Re}(\fft{Z}(\nicefrac{n}{2}, \nicefrac{n}{2}))
\end{align}
This scaling ensure that the magnitude of $Z$ is preserved:
\begin{equation}
\norm{\fftv{Z}}^2 = \abs{\fft{Z}}^2
\end{equation}
To compute the inverse of $\fftv{\cdot}$ we
undo this scaling,
undo the vectorization by filling in a subset of the elements of
$\fft{Z}$ from the vector representation,
set the other elements of $\fft{Z}$ such that Hermitian symmetry holds,
and the invert the FFT.

\section{Results}

Because our model produces a complete posterior distribution over illuminants
in the form of a covariance matrix $\boldsymbol{\Sigma}$, each of our
illuminant estimates comes with a measure of confidence in the form
of the entropy: ${1 \over 2} \log{|\boldsymbol{\Sigma}| }$ (ignoring a constant shift).
A low entropy suggests a tight concentration of the output distribution, which
tends to be well-correlated with a low error.
To demonstrate this we present a novel error metric, which is twice the area
under the curve formed by ordering all images
(the union of all test-set images from each cross-validation fold)
by ascending entropy and normalizing by the number of images.
In Figure~\ref{fig:confidence_err} we visualize this error metric and show
that our entropy-ordered error is substantially lower than the mean error
for both of our datasets, which shows that a low entropy is suggestive
of a low error.
We are not aware of any other color constancy technique which explicitly
predicts a confidence measure, and so we do not compare against any
existing technique, but it can be demonstrated that if the entropy used
to sort error is decorrelated with error (or, equivalently, if the error cannot
be sorted due to the lack of the means to sort it)
that entropy-ordered error will on average be equal to mean error.

To allow for a better understanding of our model's performance, we present
images from the Gehler-Shi dataset \cite{Gehler08,shifunt}
(Figures~\ref{fig:results1}-\ref{fig:results10})
and the Canon 1Ds MkIII camera from the Cheng \etal\, dataset \cite{Cheng14}
(Figures~\ref{fig:results11}-\ref{fig:results15}).
There results were produced using Model J presented in the main paper.
For each dataset we perform three-fold cross validation, and with that we produce
output predictions for each image along with an error measure (angular RGB
error) and an entropy measure (the entropy of the covariance matrix of our
predicted posterior distribution over illuminants). The images chosen here were
selected by sorting images from each dataset by increasing error and evenly sampling
images according to that ordering (10 from Gehler-Shi, 5 from the smaller Cheng
dataset).
This means that the first image in each sequence is the lowest error image,
and the last is the highest.
The rendered images include the color checker used in creating the ground-truth
illuminants used during training, but it should be noted that these color checkers
are masked out when these images are used during training and evaluation.
For each image we present: a) the input image, b) the predicted bivariate von Mises
distribution over illuminants, c) our estimated illuminant and white-balanced
image (produced by dividing the estimated illuminant into the input image),
and d) the ground-truth illuminant and white-balanced image.
Our log-chroma histograms are visualized using gray light de-aliasing
to assign each $(i,j)$ coordinate a color,
with a blue dot indicating the location/color of the ground-truth illuminant,
a red dot indicating our predicted illuminant $\boldsymbol{\mu}$
and a red ellipse indicating the predicted covariance of the illuminant $\boldsymbol{\Sigma}$.
The bright lines in the histogram indicate the locations where $u=0$ or $v=0$.
The reported entropy of the covariance $\boldsymbol{\Sigma}$ corresponds to the
spread of the covariance (low entropy = small spread).
We see that our low error predictions tend to
have lower entropies, and vice versa, confirming our analysis in
Figure~\ref{fig:confidence_err}.
We also see that the ground-truth
illuminant tends to lie within the estimated covariance matrix, though not
always for the largest-error images.

\begin{figure}[!]
\centering
  \begin{subfigure}[!]{1.55in}
    \includegraphics[width=1.5in]{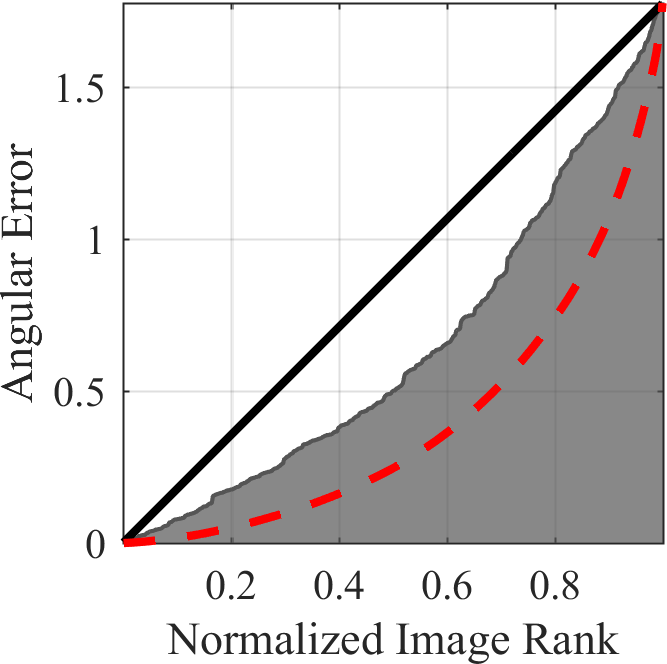}
    \caption{Gehler-Shi dataset \cite{Gehler08,shifunt}}
  \end{subfigure}
  \hspace{0.05in}
  \begin{subfigure}[!]{1.55in}
    \includegraphics[width=1.5in]{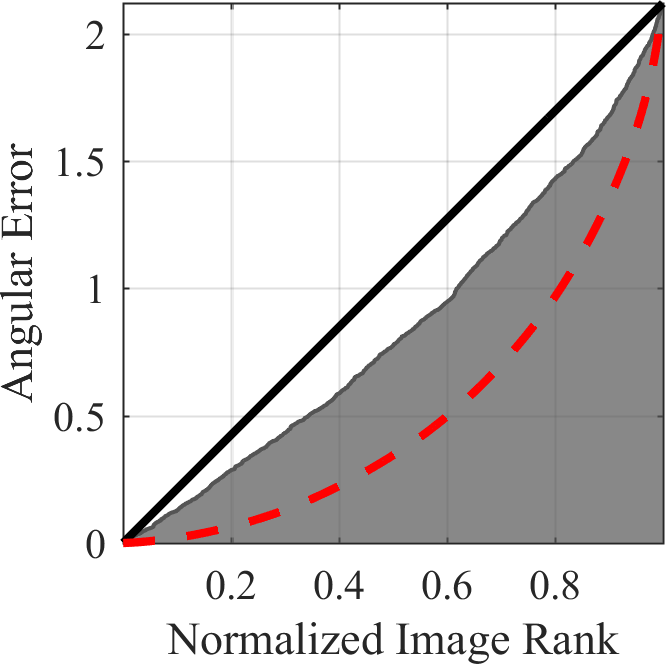}
    \caption{Cheng \etal\, dataset \cite{Cheng14}}
  \end{subfigure}
  \small
  \begin{tabular}{ r@{}c r r}
  EO Error: \quad\, & 1.287 & \hspace{1.0in} 1.696 & \hspace{0.5in}\\
  Mean Error: \quad\, & 1.775 & \hspace{1.0in} 2.121  & \hspace{0.5in}
  \end{tabular}
  \caption{
    By sorting each image by the entropy of its posterior distribution we
    can show that entropy correlates with error.
    Here we sort the images by ascending entropy and plot the cumulative sum of the error,
    filling in the area under that curve with gray.
    If entropy was not correlated with error we would expect the area under the
    curve to match the black line, and if entropy was perfectly correlated with
    error then the area under the curve would exactly match the dashed red line.
    We report twice the area under the curve as ``entropy-ordered'' error
    (mean error happens to be twice the area under the diagonal line).
    \label{fig:confidence_err}
  }
\end{figure}

In Figure~\ref{fig:realimages} we visualize a set of images taken from a Nexus
6 in the HDR+ mode \cite{Hasinoff2016} after being white-balanced by
Model Q in the main paper (the version designed to run on thumbnail images).

\section{Color Rendering}

All images are rendered by
applying the RGB gains implied by the estimated illuminant, applying some
color correction matrix (CCM)
and then applying an sRGB gamma-correction function (the
$C_{\mathrm{linear}}$ to $C_{\mathrm {srgb}}$ mapping in
\url{http://en.wikipedia.org/wiki/SRGB}).
For each camera in the datasets we use we estimate our own CCMs using the imagery,
which we present here.
These CCMs do not affect our illuminant estimation or our results,
and are only relevant to our visualizations.
Each CCM is estimated through an iterative least-squares process in which
we alternatingly:
1) estimate the ground-truth RGB gains for each image from a camera
by solving a least-squares system using our current CCM, and
2) use our current gains to estimate a row-normalized CCM using a constrained
least-squares solve.
Our estimated ground-truth gains are not used in this paper.
For the ground-truth sRGB colors of the Macbeth color chart we use the hex values
provided here: \url{http://en.wikipedia.org/wiki/ColorChecker#Colors}
which we linearize.

\begin{align}
\small{\mathrm{GehlerShi, Canon1D}}
\begin{bmatrix}
\phantom{-}2.2310 & -1.5926 & \phantom{-}0.3616 \\
-0.1494 & \phantom{-}1.4544 & -0.3050 \\
\phantom{-}0.1641 & -0.6588 & \phantom{-}1.4947 \\
\end{bmatrix} \nonumber
 \\
\small{\mathrm{GehlerShi, Canon5D}}
\begin{bmatrix}
\phantom{-}1.7494 & -0.8470 & \phantom{-}0.0976 \\
-0.1565 & \phantom{-}1.4380 & -0.2815 \\
\phantom{-}0.0786 & -0.5070 & \phantom{-}1.4284 \\
\end{bmatrix} \nonumber
\\
\small{\mathrm{Cheng, Canon1DsMkIII}}
\begin{bmatrix}
\phantom{-}1.7247 & -0.7791 & \phantom{-}0.0544 \\
-0.1436 & \phantom{-}1.4632 & -0.3195 \\
\phantom{-}0.0589 & -0.4625 & \phantom{-}1.4037 \\
\end{bmatrix} \nonumber
 \\
\small{\mathrm{Cheng, Canon600D}}
\begin{bmatrix}
\phantom{-}1.8988 & -0.9897 & \phantom{-}0.0909 \\
-0.2058 & \phantom{-}1.6396 & -0.4338 \\
\phantom{-}0.0749 & -0.7030 & \phantom{-}1.6281 \\
\end{bmatrix} \nonumber
 \\
\small{\mathrm{Cheng, FujifilmXM1}}
\begin{bmatrix}
\phantom{-}1.4183 & -0.2497 & -0.1686 \\
-0.2230 & \phantom{-}1.6449 & -0.4219 \\
\phantom{-}0.0785 & -0.5980 & \phantom{-}1.5195 \\
\end{bmatrix} \nonumber
 \\
\small{\mathrm{Cheng, NikonD5200}}
\begin{bmatrix}
\phantom{-}1.3792 & -0.3134 & -0.0659 \\
-0.0826 & \phantom{-}1.3759 & -0.2932 \\
\phantom{-}0.0483 & -0.4553 & \phantom{-}1.4070 \\
\end{bmatrix} \nonumber
 \\
\small{\mathrm{Cheng, OlympusEPL6}}
\begin{bmatrix}
\phantom{-}1.6565 & -0.4971 & -0.1595 \\
-0.3335 & \phantom{-}1.7772 & -0.4437 \\
\phantom{-}0.0895 & -0.7023 & \phantom{-}1.6128 \\
\end{bmatrix} \nonumber
 \\
\small{\mathrm{Cheng, PanasonicGX1}}
\begin{bmatrix}
\phantom{-}1.5629 & -0.5117 & -0.0512 \\
-0.2472 & \phantom{-}1.7590 & -0.5117 \\
\phantom{-}0.1395 & -0.8945 & \phantom{-}1.7550 \\
\end{bmatrix} \nonumber
 \\
\small{\mathrm{Cheng, SamsungNX2000}}
\begin{bmatrix}
\phantom{-}1.5770 & -0.4351 & -0.1419 \\
-0.1747 & \phantom{-}1.5225 & -0.3477 \\
\phantom{-}0.0573 & -0.6397 & \phantom{-}1.5825 \\
\end{bmatrix} \nonumber
 \\
\small{\mathrm{Cheng, SonyA57}}
\begin{bmatrix}
\phantom{-}1.5963 & -0.5545 & -0.0418 \\
-0.1343 & \phantom{-}1.5331 & -0.3988 \\
\phantom{-}0.0563 & -0.4026 & \phantom{-}1.3463 \\
\end{bmatrix} \nonumber
\end{align}

\begin{figure*}[!]
\centering
  \begin{subfigure}[!]{1.7in}
    \includegraphics[width=1.6in]{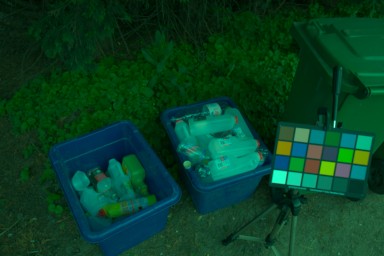}
    \caption{\footnotesize Input Image}
  \end{subfigure}
  \begin{subfigure}[!]{1.17in}
    \includegraphics[width=1.07in]{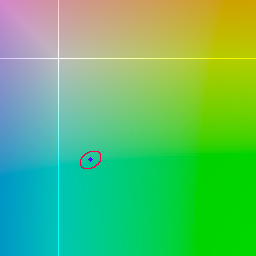}
    \caption{\footnotesize Illuminant Posterior}
  \end{subfigure}
\begin{subfigure}[!]{1.9in}
    \includegraphics[width=0.133in]{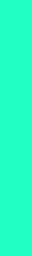}\!
    \includegraphics[width=1.6in]{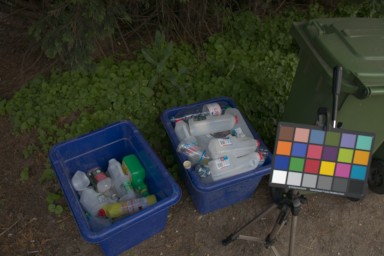}
    \caption{\footnotesize Our prediction}
  \end{subfigure}
  \begin{subfigure}[!]{1.9in}
    \includegraphics[width=0.133in]{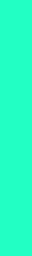}\!
    \includegraphics[width=1.6in]{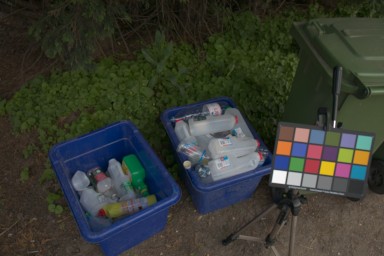}
    \caption{\footnotesize Ground Truth}
  \end{subfigure}
  \caption{
    A result from the Gehler-Shi dataset using Model J. Error = $0.02$\textdegree, entropy = $-6.48$
    \label{fig:results1}
  }
\end{figure*}

\begin{figure*}[!]
\centering
  \begin{subfigure}[!]{1.7in}
    \includegraphics[width=1.6in]{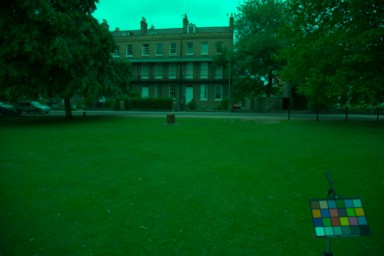}
    \caption{\footnotesize Input Image}
  \end{subfigure}
  \begin{subfigure}[!]{1.17in}
    \includegraphics[width=1.07in]{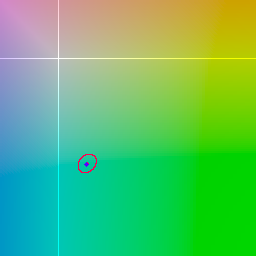}
    \caption{\footnotesize Illuminant Posterior}
  \end{subfigure}
\begin{subfigure}[!]{1.9in}
    \includegraphics[width=0.133in]{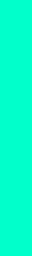}\!
    \includegraphics[width=1.6in]{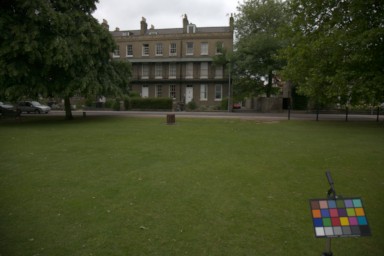}
    \caption{\footnotesize Our prediction}
  \end{subfigure}
  \begin{subfigure}[!]{1.9in}
    \includegraphics[width=0.133in]{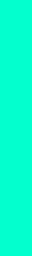}\!
    \includegraphics[width=1.6in]{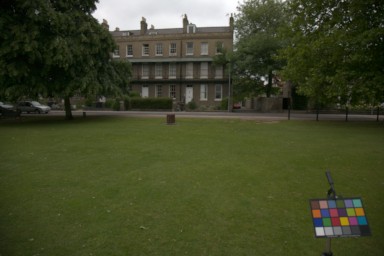}
    \caption{\footnotesize Ground Truth}
  \end{subfigure}
  \caption{
    A result from the Gehler-Shi dataset using Model J. Error = $0.26$\textdegree, entropy = $-6.55$
    \label{fig:results2}
  }
\end{figure*}

\begin{figure*}[!]
\centering
  \begin{subfigure}[!]{1.7in}
    \includegraphics[width=1.6in]{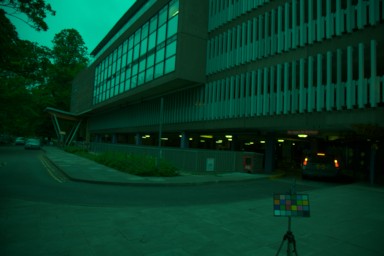}
    \caption{\footnotesize Input Image}
  \end{subfigure}
  \begin{subfigure}[!]{1.17in}
    \includegraphics[width=1.07in]{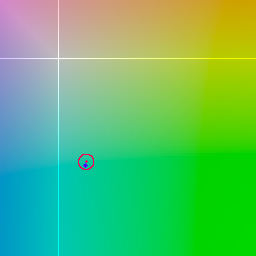}
    \caption{\footnotesize Illuminant Posterior}
  \end{subfigure}
\begin{subfigure}[!]{1.9in}
    \includegraphics[width=0.133in]{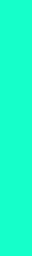}\!
    \includegraphics[width=1.6in]{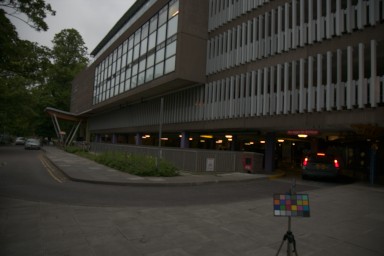}
    \caption{\footnotesize Our prediction}
  \end{subfigure}
  \begin{subfigure}[!]{1.9in}
    \includegraphics[width=0.133in]{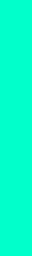}\!
    \includegraphics[width=1.6in]{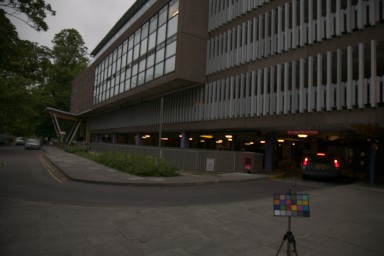}
    \caption{\footnotesize Ground Truth}
  \end{subfigure}
  \caption{
    A result from the Gehler-Shi dataset using Model J. Error = $0.46$\textdegree, entropy = $-6.91$
    \label{fig:results3}
  }
\end{figure*}

\begin{figure*}[!]
\centering
  \begin{subfigure}[!]{1.7in}
    \includegraphics[width=1.6in]{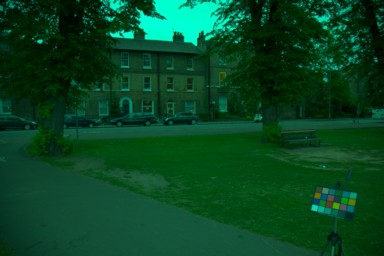}
    \caption{\footnotesize Input Image}
  \end{subfigure}
  \begin{subfigure}[!]{1.17in}
    \includegraphics[width=1.07in]{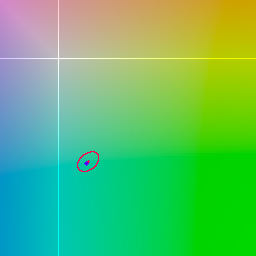}
    \caption{\footnotesize Illuminant Posterior}
  \end{subfigure}
\begin{subfigure}[!]{1.9in}
    \includegraphics[width=0.133in]{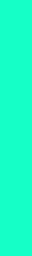}\!
    \includegraphics[width=1.6in]{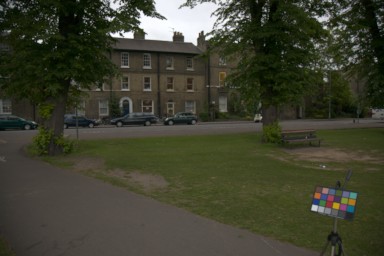}
    \caption{\footnotesize Our prediction}
  \end{subfigure}
  \begin{subfigure}[!]{1.9in}
    \includegraphics[width=0.133in]{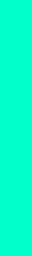}\!
    \includegraphics[width=1.6in]{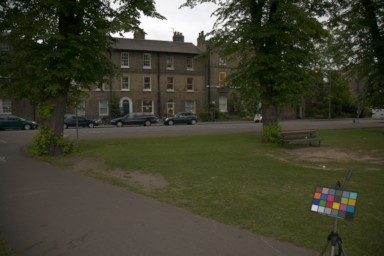}
    \caption{\footnotesize Ground Truth}
  \end{subfigure}
  \caption{
    A result from the Gehler-Shi dataset using Model J. Error = $0.63$\textdegree, entropy = $-6.37$
    \label{fig:results4}
  }
\end{figure*}

\begin{figure*}[!]
\centering
  \begin{subfigure}[!]{1.7in}
    \includegraphics[width=1.6in]{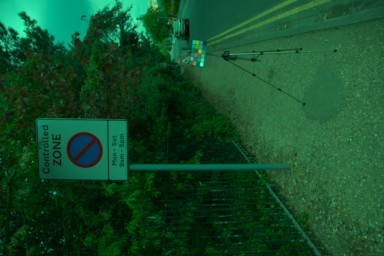}
    \caption{\footnotesize Input Image}
  \end{subfigure}
  \begin{subfigure}[!]{1.17in}
    \includegraphics[width=1.07in]{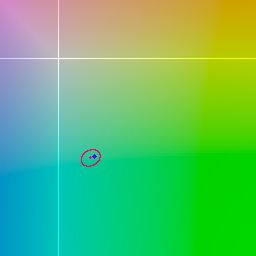}
    \caption{\footnotesize Illuminant Posterior}
  \end{subfigure}
\begin{subfigure}[!]{1.9in}
    \includegraphics[width=0.133in]{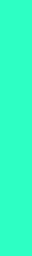}\!
    \includegraphics[width=1.6in]{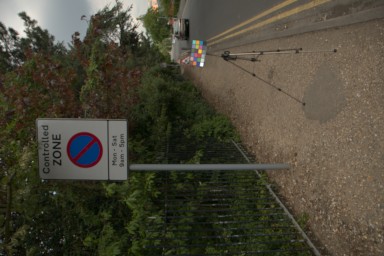}
    \caption{\footnotesize Our prediction}
  \end{subfigure}
  \begin{subfigure}[!]{1.9in}
    \includegraphics[width=0.133in]{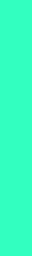}\!
    \includegraphics[width=1.6in]{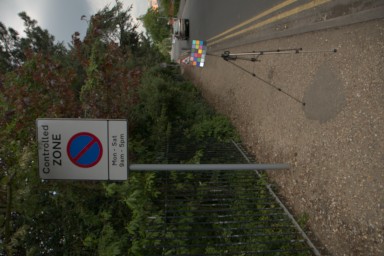}
    \caption{\footnotesize Ground Truth}
  \end{subfigure}
  \caption{
    A result from the Gehler-Shi dataset using Model J. Error = $0.83$\textdegree, entropy = $-6.62$
    \label{fig:results5}
  }
\end{figure*}

\begin{figure*}[!]
\centering
  \begin{subfigure}[!]{1.7in}
    \includegraphics[width=1.6in]{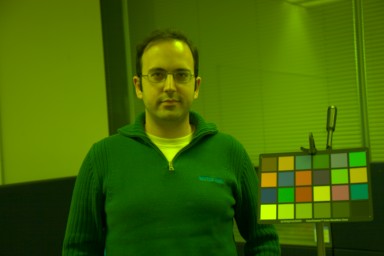}
    \caption{\footnotesize Input Image}
  \end{subfigure}
  \begin{subfigure}[!]{1.17in}
    \includegraphics[width=1.07in]{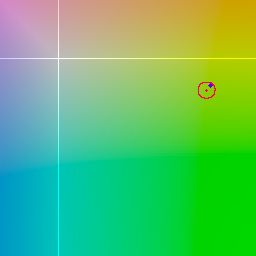}
    \caption{\footnotesize Illuminant Posterior}
  \end{subfigure}
\begin{subfigure}[!]{1.9in}
    \includegraphics[width=0.133in]{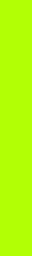}\!
    \includegraphics[width=1.6in]{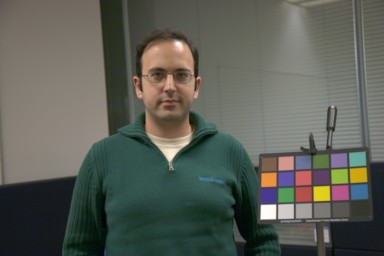}
    \caption{\footnotesize Our prediction}
  \end{subfigure}
  \begin{subfigure}[!]{1.9in}
    \includegraphics[width=0.133in]{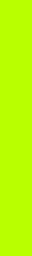}\!
    \includegraphics[width=1.6in]{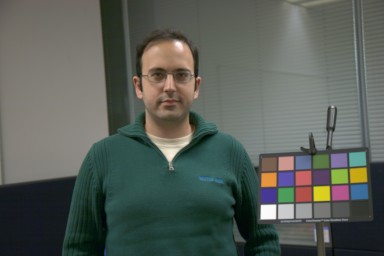}
    \caption{\footnotesize Ground Truth}
  \end{subfigure}
  \caption{
    A result from the Gehler-Shi dataset using Model J. Error = $1.19$\textdegree, entropy = $-6.71$
    \label{fig:results6}
  }
\end{figure*}

\begin{figure*}[!]
\centering
  \begin{subfigure}[!]{1.7in}
    \includegraphics[width=1.6in]{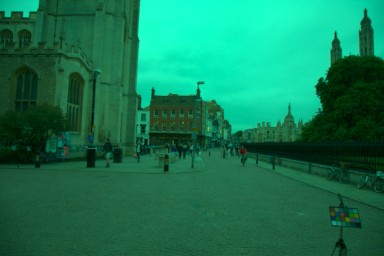}
    \caption{\footnotesize Input Image}
  \end{subfigure}
  \begin{subfigure}[!]{1.17in}
    \includegraphics[width=1.07in]{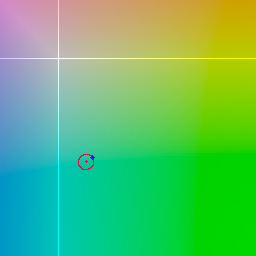}
    \caption{\footnotesize Illuminant Posterior}
  \end{subfigure}
\begin{subfigure}[!]{1.9in}
    \includegraphics[width=0.133in]{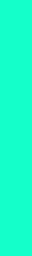}\!
    \includegraphics[width=1.6in]{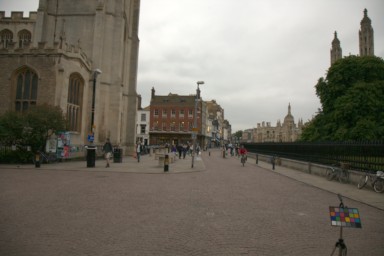}
    \caption{\footnotesize Our prediction}
  \end{subfigure}
  \begin{subfigure}[!]{1.9in}
    \includegraphics[width=0.133in]{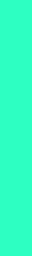}\!
    \includegraphics[width=1.6in]{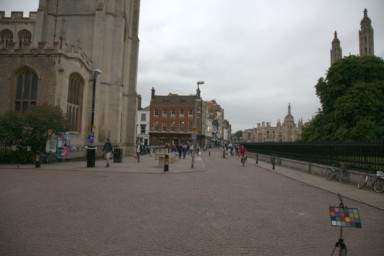}
    \caption{\footnotesize Ground Truth}
  \end{subfigure}
  \caption{
    A result from the Gehler-Shi dataset using Model J. Error = $1.61$\textdegree, entropy = $-6.88$
    \label{fig:results7}
  }
\end{figure*}

\begin{figure*}[!]
\centering
  \begin{subfigure}[!]{1.7in}
    \includegraphics[width=1.6in]{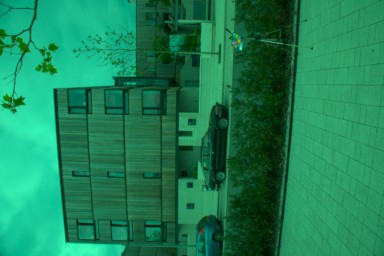}
    \caption{\footnotesize Input Image}
  \end{subfigure}
  \begin{subfigure}[!]{1.17in}
    \includegraphics[width=1.07in]{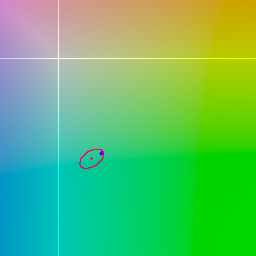}
    \caption{\footnotesize Illuminant Posterior}
  \end{subfigure}
\begin{subfigure}[!]{1.9in}
    \includegraphics[width=0.133in]{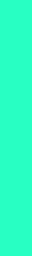}\!
    \includegraphics[width=1.6in]{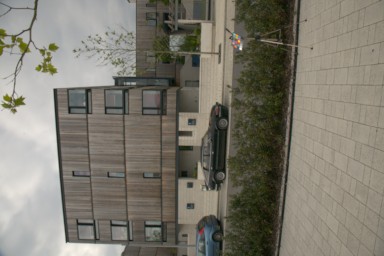}
    \caption{\footnotesize Our prediction}
  \end{subfigure}
  \begin{subfigure}[!]{1.9in}
    \includegraphics[width=0.133in]{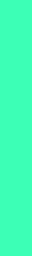}\!
    \includegraphics[width=1.6in]{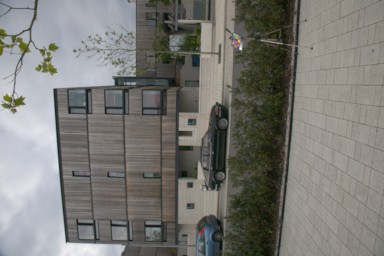}
    \caption{\footnotesize Ground Truth}
  \end{subfigure}
  \caption{
    A result from the Gehler-Shi dataset using Model J. Error = $2.35$\textdegree, entropy = $-6.32$
    \label{fig:results8}
  }
\end{figure*}

\begin{figure*}[!]
\centering
  \begin{subfigure}[!]{1.7in}
    \includegraphics[width=1.6in]{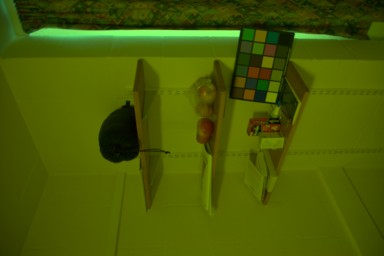}
    \caption{\footnotesize Input Image}
  \end{subfigure}
  \begin{subfigure}[!]{1.17in}
    \includegraphics[width=1.07in]{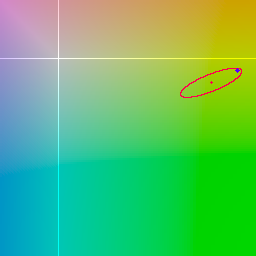}
    \caption{\footnotesize Illuminant Posterior}
  \end{subfigure}
\begin{subfigure}[!]{1.9in}
    \includegraphics[width=0.133in]{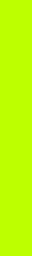}\!
    \includegraphics[width=1.6in]{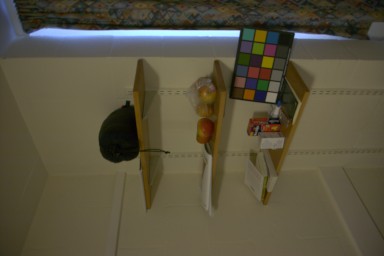}
    \caption{\footnotesize Our prediction}
  \end{subfigure}
  \begin{subfigure}[!]{1.9in}
    \includegraphics[width=0.133in]{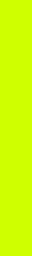}\!
    \includegraphics[width=1.6in]{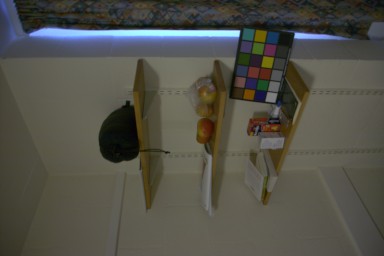}
    \caption{\footnotesize Ground Truth}
  \end{subfigure}
  \caption{
    A result from the Gehler-Shi dataset using Model J. Error = $3.84$\textdegree, entropy = $-5.28$
    \label{fig:results9}
  }
\end{figure*}

\begin{figure*}[!]
\centering
  \begin{subfigure}[!]{1.7in}
    \includegraphics[width=1.6in]{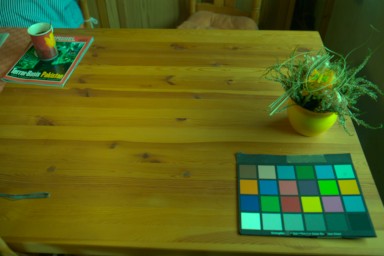}
    \caption{\footnotesize Input Image}
  \end{subfigure}
  \begin{subfigure}[!]{1.17in}
    \includegraphics[width=1.07in]{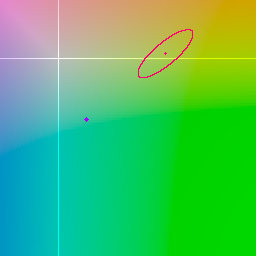}
    \caption{\footnotesize Illuminant Posterior}
  \end{subfigure}
\begin{subfigure}[!]{1.9in}
    \includegraphics[width=0.133in]{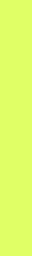}\!
    \includegraphics[width=1.6in]{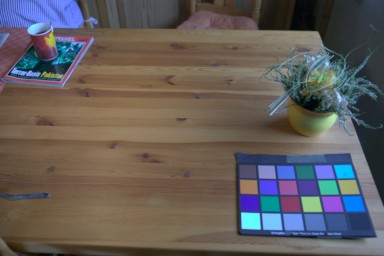}
    \caption{\footnotesize Our prediction}
  \end{subfigure}
  \begin{subfigure}[!]{1.9in}
    \includegraphics[width=0.133in]{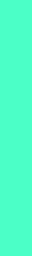}\!
    \includegraphics[width=1.6in]{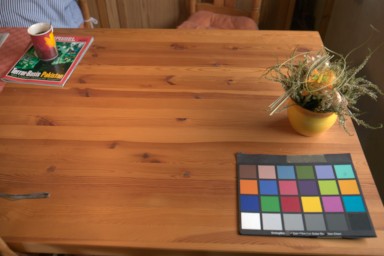}
    \caption{\footnotesize Ground Truth}
  \end{subfigure}
  \caption{
    A result from the Gehler-Shi dataset using Model J. Error = $21.64$\textdegree, entropy = $-4.95$
    \label{fig:results10}
  }
\end{figure*}

\begin{figure*}[!]
\centering
  \begin{subfigure}[!]{1.7in}
    \includegraphics[width=1.6in]{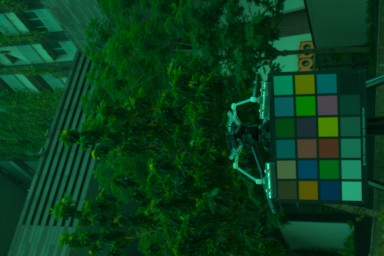}
    \caption{\footnotesize Input Image}
  \end{subfigure}
  \begin{subfigure}[!]{1.17in}
    \includegraphics[width=1.07in]{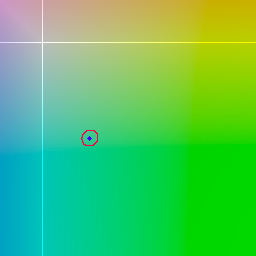}
    \caption{\footnotesize Illuminant Posterior}
  \end{subfigure}
\begin{subfigure}[!]{1.9in}
    \includegraphics[width=0.133in]{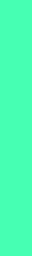}\!
    \includegraphics[width=1.6in]{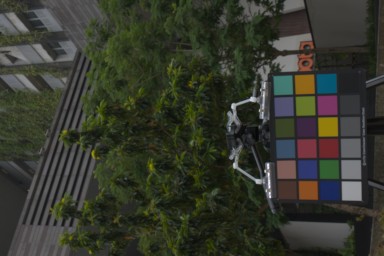}
    \caption{\footnotesize Our prediction}
  \end{subfigure}
  \begin{subfigure}[!]{1.9in}
    \includegraphics[width=0.133in]{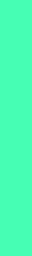}\!
    \includegraphics[width=1.6in]{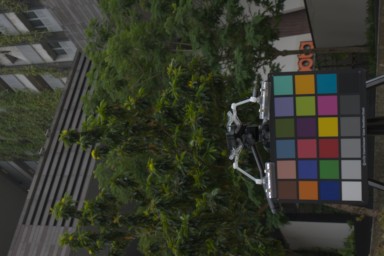}
    \caption{\footnotesize Ground Truth}
  \end{subfigure}
  \caption{
    A result from the Cheng dataset using Model J. Error = $0.12$\textdegree, entropy = $-6.82$
    \label{fig:results11}
  }
\end{figure*}

\begin{figure*}[!]
\centering
  \begin{subfigure}[!]{1.7in}
    \includegraphics[width=1.6in]{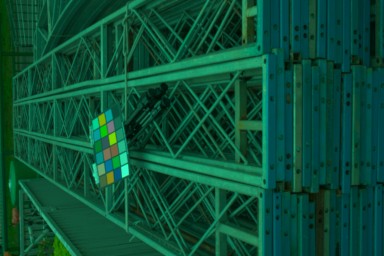}
    \caption{\footnotesize Input Image}
  \end{subfigure}
  \begin{subfigure}[!]{1.17in}
    \includegraphics[width=1.07in]{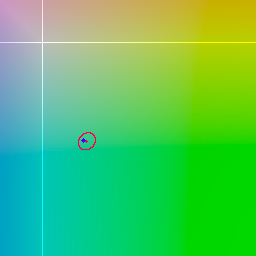}
    \caption{\footnotesize Illuminant Posterior}
  \end{subfigure}
\begin{subfigure}[!]{1.9in}
    \includegraphics[width=0.133in]{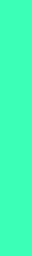}\!
    \includegraphics[width=1.6in]{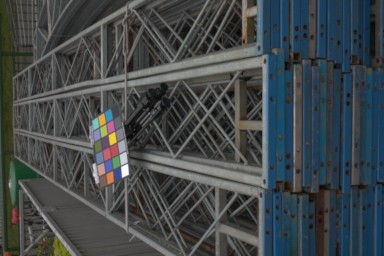}
    \caption{\footnotesize Our prediction}
  \end{subfigure}
  \begin{subfigure}[!]{1.9in}
    \includegraphics[width=0.133in]{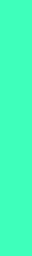}\!
    \includegraphics[width=1.6in]{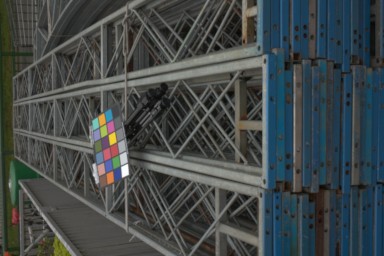}
    \caption{\footnotesize Ground Truth}
  \end{subfigure}
  \caption{
    A result from the Cheng dataset using Model J. Error = $0.64$\textdegree, entropy = $-6.69$
    \label{fig:results12}
  }
\end{figure*}

\begin{figure*}[!]
\centering
  \begin{subfigure}[!]{1.7in}
    \includegraphics[width=1.6in]{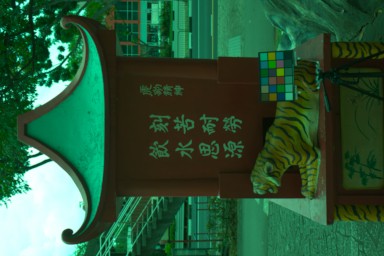}
    \caption{\footnotesize Input Image}
  \end{subfigure}
  \begin{subfigure}[!]{1.17in}
    \includegraphics[width=1.07in]{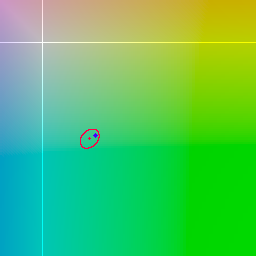}
    \caption{\footnotesize Illuminant Posterior}
  \end{subfigure}
\begin{subfigure}[!]{1.9in}
    \includegraphics[width=0.133in]{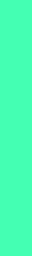}\!
    \includegraphics[width=1.6in]{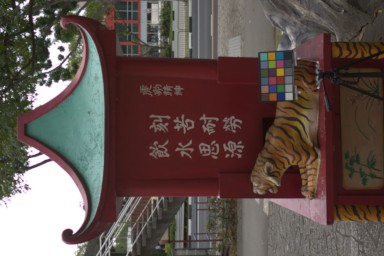}
    \caption{\footnotesize Our prediction}
  \end{subfigure}
  \begin{subfigure}[!]{1.9in}
    \includegraphics[width=0.133in]{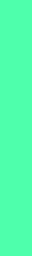}\!
    \includegraphics[width=1.6in]{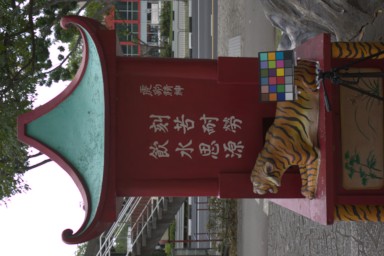}
    \caption{\footnotesize Ground Truth}
  \end{subfigure}
  \caption{
    A result from the Cheng dataset using Model J. Error = $1.37$\textdegree, entropy = $-6.48$
    \label{fig:results13}
  }
\end{figure*}

\begin{figure*}[!]
\centering
  \begin{subfigure}[!]{1.7in}
    \includegraphics[width=1.6in]{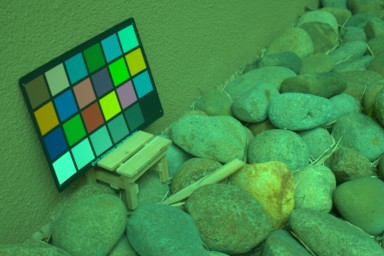}
    \caption{\footnotesize Input Image}
  \end{subfigure}
  \begin{subfigure}[!]{1.17in}
    \includegraphics[width=1.07in]{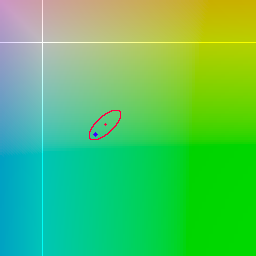}
    \caption{\footnotesize Illuminant Posterior}
  \end{subfigure}
\begin{subfigure}[!]{1.9in}
    \includegraphics[width=0.133in]{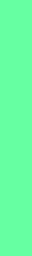}\!
    \includegraphics[width=1.6in]{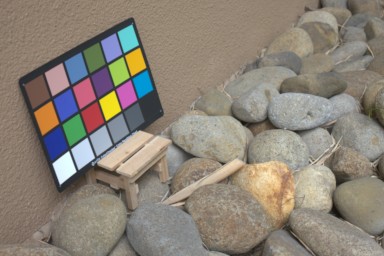}
    \caption{\footnotesize Our prediction}
  \end{subfigure}
  \begin{subfigure}[!]{1.9in}
    \includegraphics[width=0.133in]{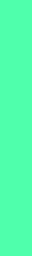}\!
    \includegraphics[width=1.6in]{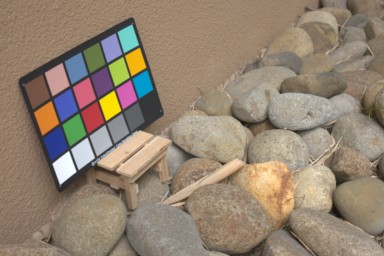}
    \caption{\footnotesize Ground Truth}
  \end{subfigure}
  \caption{
    A result from the Cheng dataset using Model J. Error = $2.69$\textdegree, entropy = $-5.82$
    \label{fig:results14}
  }
\end{figure*}

\begin{figure*}[!]
\centering
  \begin{subfigure}[!]{1.7in}
    \includegraphics[width=1.6in]{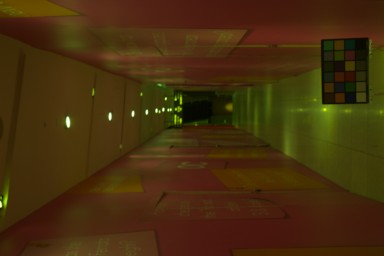}
    \caption{\footnotesize Input Image}
  \end{subfigure}
  \begin{subfigure}[!]{1.17in}
    \includegraphics[width=1.07in]{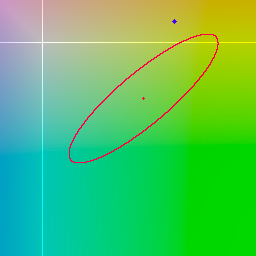}
    \caption{\footnotesize Illuminant Posterior}
  \end{subfigure}
\begin{subfigure}[!]{1.9in}
    \includegraphics[width=0.133in]{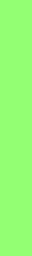}\!
    \includegraphics[width=1.6in]{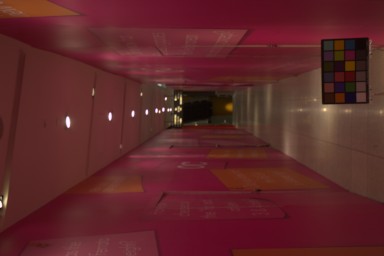}
    \caption{\footnotesize Our prediction}
  \end{subfigure}
  \begin{subfigure}[!]{1.9in}
    \includegraphics[width=0.133in]{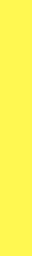}\!
    \includegraphics[width=1.6in]{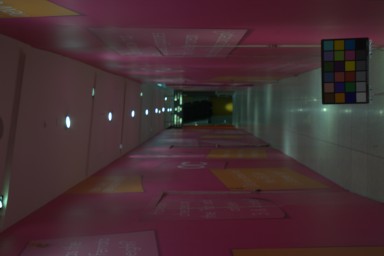}
    \caption{\footnotesize Ground Truth}
  \end{subfigure}
  \caption{
    A result from the Cheng dataset using Model J. Error = $17.85$\textdegree, entropy = $-3.04$
    \label{fig:results15}
  }
\end{figure*}

\begin{figure*}[!]
\centering
  \includegraphics[width=\threewidth]{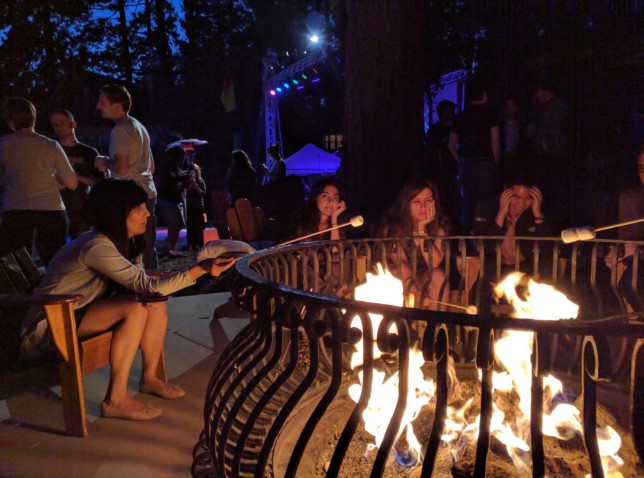}
  \includegraphics[width=\threewidth]{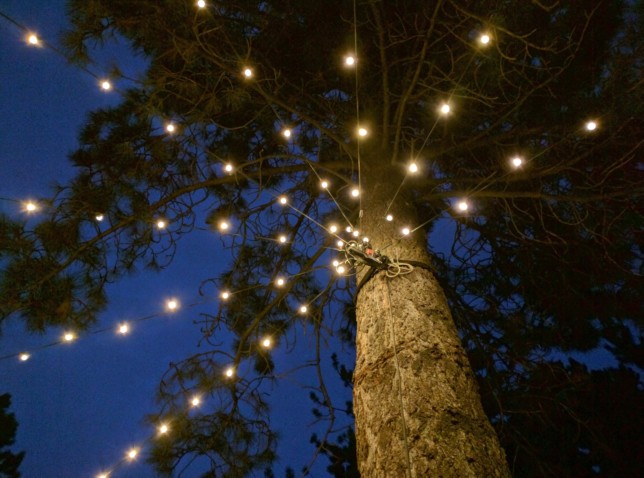}
  \includegraphics[width=\threewidth]{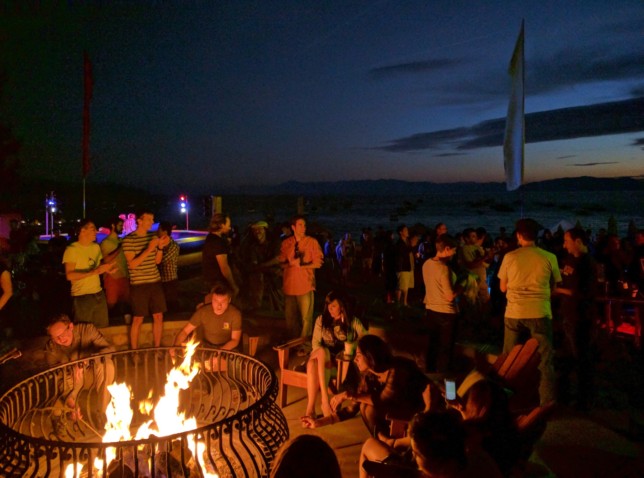}
  \includegraphics[width=\threewidth]{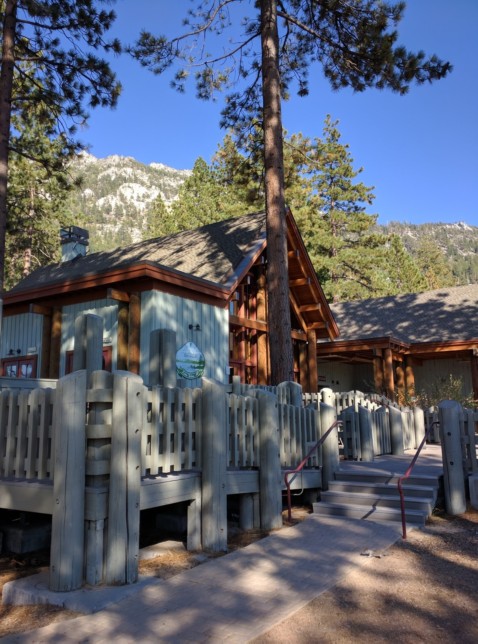}
  \includegraphics[width=\threewidth]{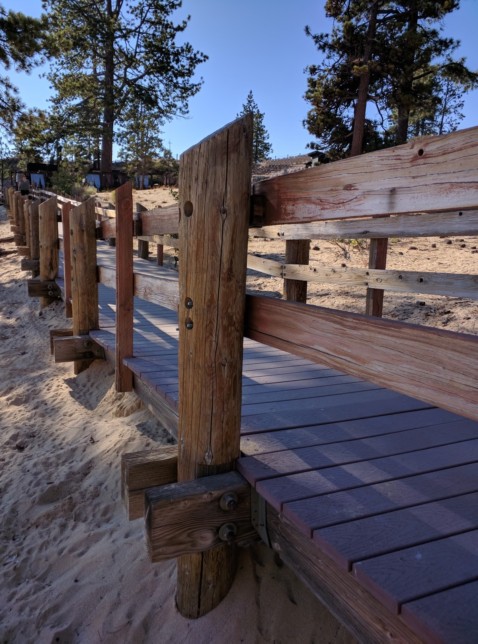}
  \includegraphics[width=\threewidth]{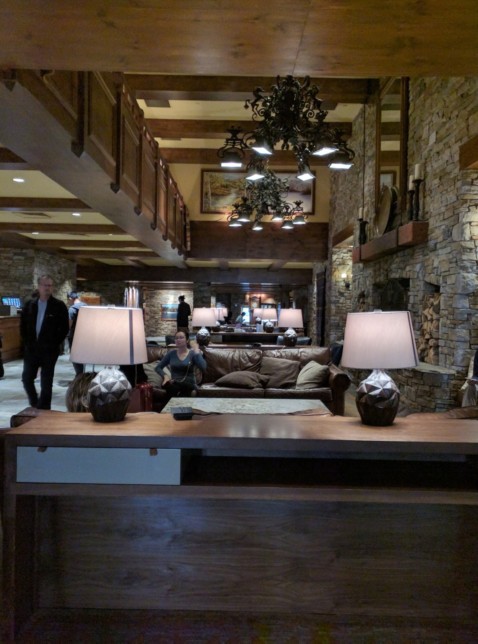}
  \includegraphics[width=\threewidth]{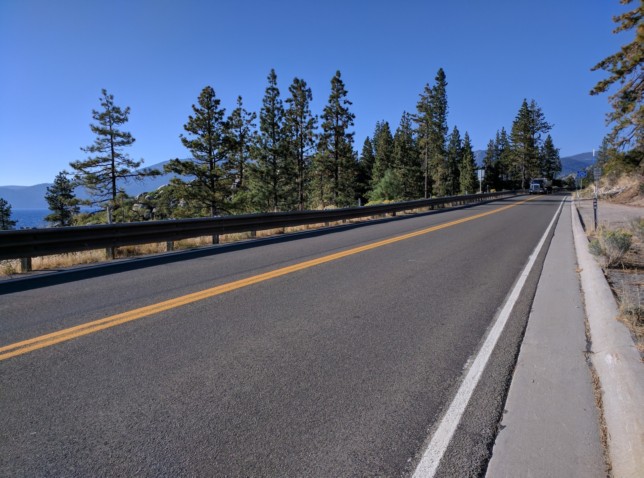}
  \includegraphics[width=\threewidth]{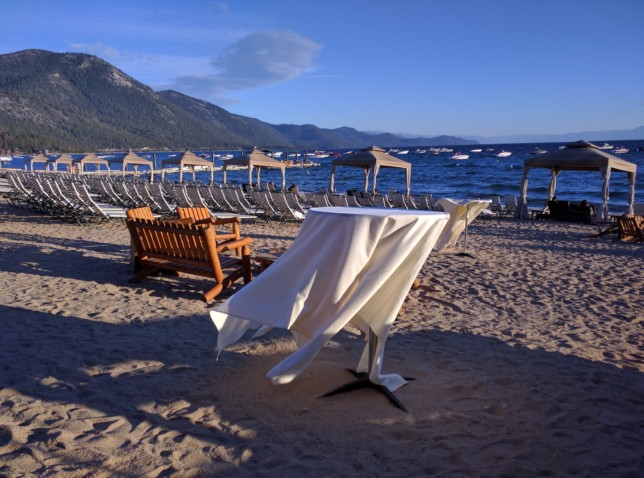}
  \includegraphics[width=\threewidth]{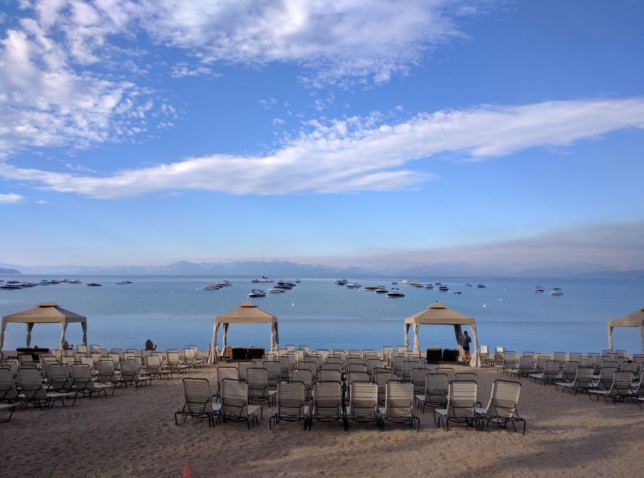}
  \caption{
    A sampling of unedited HDR+\cite{Hasinoff2016} images from a Nexus 6, after
    being processed with Model Q.
    \label{fig:realimages}
  }
\end{figure*}

\end{document}